\documentclass[letterpaper]{article} 
\usepackage{aaai25}  
\usepackage{times}  
\usepackage{helvet}  
\usepackage{courier}  
\usepackage[hyphens]{url}  
\usepackage{graphicx} 
\usepackage{natbib}  
\usepackage{caption} 
\usepackage{algorithm}
\usepackage{algorithmic}
\usepackage{subcaption}
\usepackage{multirow}
\usepackage{array}
\usepackage{rotating}
\usepackage{booktabs}
\usepackage{pdfpages}
\usepackage{graphicx}
\usepackage{xcolor}
\newcommand{\answerYes}[1]{\textcolor{blue}{#1}} 
\newcommand{\answerNo}[1]{\textcolor{teal}{#1}} 
\newcommand{\answerNA}[1]{\textcolor{gray}{#1}}

\usepackage{newfloat}
\usepackage{listings}
\DeclareCaptionStyle{ruled}{labelfont=normalfont,labelsep=colon,strut=off} 
\floatstyle{ruled}
\newfloat{listing}{tb}{lst}{}
\floatname{listing}{Listing}

\title{Misleading through Inconsistency: A Benchmark for Political Inconsistencies Detection}
\author {
    Nursulu Sagimbayeva\textsuperscript{\rm 1},
    Ruveyda Betül Bahçeci\textsuperscript{\rm 2},
    Ingmar Weber\textsuperscript{\rm 1}
}
\affiliations {
    \textsuperscript{\rm 1}Saarland Informatics Campus, Saarland University\\
    \textsuperscript{\rm 2}Saarland University\\
    nusa00001@uni-saarland.de, ruba00002@uni-saarland.de, iweber@cs.uni-saarland.de
}

\usepackage{bibentry}

\begin{document}

\maketitle

\begin{abstract}
Inconsistent political statements represent a form of misinformation. They erode public trust and pose challenges to accountability, when left unnoticed.
Detecting inconsistencies automatically could support journalists in asking clarification questions, thereby helping to keep politicians accountable. We propose the Inconsistency detection task and develop a scale of inconsistency types to prompt NLP-research in this direction.  To provide a resource for detecting inconsistencies in a political domain, we present a dataset of 698 human-annotated pairs of political statements with explanations of the annotators' reasoning for 237 samples. The statements mainly come from voting assistant platforms such as Wahl-O-Mat in Germany and Smartvote in Switzerland, reflecting real-world political issues. We benchmark Large Language Models (LLMs) on our dataset and show that in general, they are as good as humans at detecting inconsistencies, and might be even better than individual humans at predicting the crowd-annotated ground-truth. However, when it comes to identifying fine-grained inconsistency types, none of the model have reached the upper bound of performance (due to natural labeling variation), thus leaving room for improvement. We make our dataset and code publicly available.\footnote{\url{https://github.com/nursaltyn/Inconsistency_Detection_Benchmark}}
\end{abstract}

\section{Introduction}

Once elected, politicians have an unspoken duty to fulfill their campaign promises. While consistency between stated beliefs and legislative actions is associated with credibility and ideological commitment, inconsistencies can undermine public trust and support \citep{Friedman_Kampf_2020}. 
\begin{figure}[ht!]
  \centering
  \includegraphics[width=\linewidth]{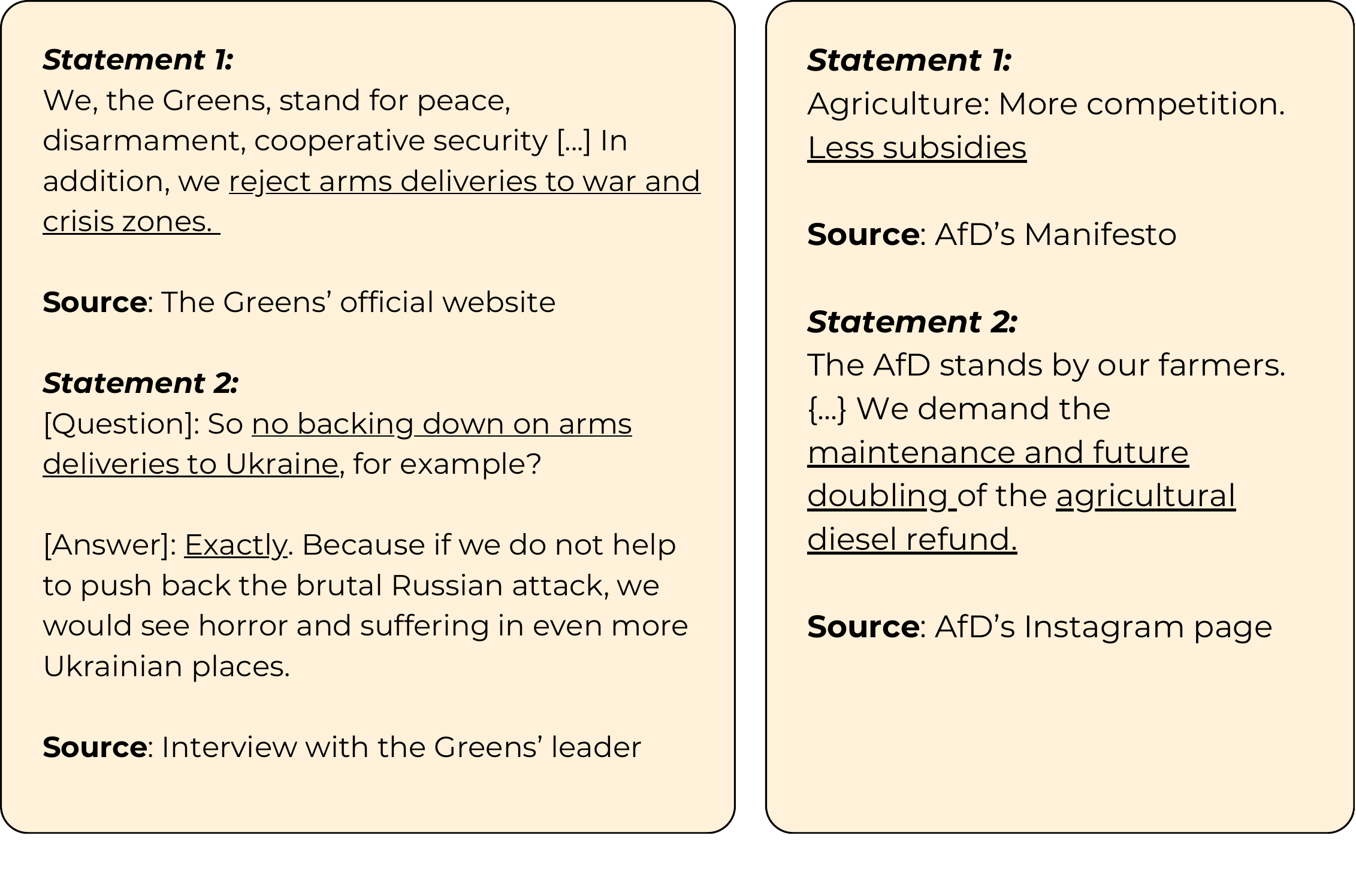}  
  \caption{Example of inconsistencies from the Green party (left-wing) and the AfD party (far-right), Germany. Underlined are spans that explain the inconsistencies. Original statements can be found in Appendix \ref{orig_statements_politics}.}
  \label{fig:statement_2}
\end{figure}

Figure \ref{fig:statement_2} presents two recent examples of inconsistent statements in German politics. One instance is the Green Party, which abandoned its long-standing opposition to arms exports following Russia's 2022 invasion of Ukraine. Likewise, the AfD, initially opposed to agricultural subsidies, later called for "doubling the diesel refund" in response to large-scale farmer protests, effectively endorsing a form of subsidy. Both cases received significant media attention, highlighting the impact of perceived inconsistencies in public discourse.

\textbf{Challenging politicians about such inconsistencies is a common tactic for journalists} during press conferences to expose contradictions and prompt clarifications. However, with the massive amount of digital content produced by parties and their members, it becomes more challenging to detect inconsistencies manually. Moreover, parties use different platforms to communicate with their electorate, and can alter their messages based on the platform: for example, they can express more aggressive views on TikTok and more moderate on Facebook \citep{tiktok_agression}. These challenges highlight the need for automated systems to detect inconsistencies.

\textbf{Inconsistency Detection in political statements }extends Natural Language Inference (NLI) task, also known as Recognizing Textual Entailment (RTE). \citep{MacCartney_2009, Dagan_Glickman_2004}. While NLI classifies the relationship between two texts as Entailment, Unrelated, or Contradiction, political inconsistency detection involves nuances not fully captured by traditional NLI taxonomies. 
For example, consider the following statements:

\textbf{A:} Family is the foundation of society and should \underline{enjoy special protection and value}.

\textbf{B:} Parents should \underline{bear the costs} of looking after their children in daycare centers.

At a first glance, $A$ and $B$ might look unrelated, since there is no direct contradiction between them. However, ideologically they go in opposite directions with respect to family support. One could expect that, everything else being equal, a consistent politician who claims that "family should enjoy special protection and value" will also support free daycare instead of insisting that parents bear the costs. We compare Inconsistency detection to other NLP tasks in Section \ref{sec:related_work_comparison}.

In recent years, Large Language Models (LLMs) have shown remarkable capabilities across various NLP tasks \citep{Wei2022EmergentAO, Zhao2023ASO}, suggesting the potential for deploying Inconsistency detection in real-world scenarios. However, to our knowledge, there are no resources focused on inconsistencies in a political domain. \textbf{To address this gap, we introduce a dataset }comprising 698 human-annotated political statements and 334 human explanations for 237 samples (Section \ref{dataset}). The inter-annotator agreement measured by Krippendorff’s alpha is \textbf{0.528} for the 5-class setting (ordinal scale) and \textbf{0.507} for the 3-class setting (nominal scale), which reflects the inherent subjectivity of the task. We evaluate several LLMs on our benchmark and estimate an upper bound of performance for the Inconsistency detection task due to the inherent labeling noise (Section \ref{sec:model_eval}). 

It is important to emphasize that changing views is not inherently bad and is often unavoidable. Politicians constantly face pressure to accommodate different audiences and react to changes in the environment, which can lead to inconsistencies between their messaging and policies \citep{Friedman_Kampf_2020}. In their worst, however, inconsistencies might be a sign of populism
\citep{Frisell2006Populism, Alesina_1988}, or even deliberate deception and manipulation. Therefore, Inconsistency detection systems have the potential to promote transparency and accountability.

\section{Task formulation}
\label{task_formulation}

We define our task as follows: given a pair of statements $A$ and $B$, detect the relationship between them as one of the labels: \{\textit{Unrelated}, \textit{Consistent}, \textit{Inconsistent}\}. If the class is Inconsistent, detect a subtype of inconsistency: \{\textit{Surface contradiction}, \textit{Factual inconsistency}, \textit{Indirect (Value) inconsistency}\}\footnote{We also considered adding Stereotypes/Expectation violation category, but refrained from it. See details in Appendix \ref{stereotypes_category}}. Under our settings:
\begin{itemize}
    \item Statements $A$ and $B$ can be of arbitrary length (from one-liner social media posts to several-page manifestos). 
    \item Statements $A$ and $B$ might be a part of the bigger document $D$, in which case self-inconsistency detection is performed for $D$.
    \item To simplify the task, we assume that $A$ and $B$ were said by the same actor on the same day. This is important because, in reality, time and context significantly influence perceptions of inconsistency. For example, promises made before unexpected crises, such as a pandemic, may become unfulfillable due to changing circumstances, and voters may not view this as inconsistent.
\end{itemize}

\begin{table*}[ht]
\centering
\small
\begin{tabular} 
{@{}p{2cm}p{3.5cm}p{3cm}p{6cm}@{}}
\toprule
\textbf{Type} & \textbf{Description} & \textbf{Example} & \textbf{Explanation} \\ \midrule
Surface contradiction & The strongest degree of inconsistency. No external or specialized knowledge is required to detect it. Understanding logical form and/or language in $A$ and $B$ alone is enough. If $A$ is True, $B$ must be False, and vice versa & 

Ex. 1) \textbf{A:} All kikis are bobable. 

\textbf{B:} This kiki is not bobable.

Ex. 2) \textbf{A:} We \textit{support} the yellow party.

\textbf{B:} We \textit{never want to collaborate} with the yellow party.
& Ex. 1) We don't have to know what "kiki" and "bobable" mean in the real world. The logical form is: A: All X are Y, B: This Y is not X, and these are mutually exclusive.

Ex. 2) Here, we don't have to know who the speaker is, and whether the yellow party exists. From the linguistic formulation only, we can judge that the statements are contradictory.

\\ \midrule
Factual inconsistency & Having external knowledge about the world \textit{beyond} what is said in $A$ and $B$ is required. This can include laws of physics, economics principles, etc., as well as real-world events and evidence. If $A$ is True, it directly challenges the Truth of $B$, but does not necessarily make $B$ impossible, and vice versa. & \textbf{A:} We will provide extensive social benefits.

\textbf{B:} We will lower all taxes.    
 & Based on empirical evidence, increasing social benefits usually requires higher taxes. This understanding is necessary to detect inconsistency. At the same time, $A$ and $B$ are not mutually exclusive - the government might also take on more debt or use other ways to increase social benefits. However, based only on the information we have and empirical evidence, we can assume that $A$ and $B$ are Factually inconsistent. \\ \midrule
Indirect inconsistency (Value inconsistency) & If $A$ is True, it \textit{doesn't} directly challenge the Truth of $B$, and vice versa. However, $A$ and $B$ go in \textit{opposite} directions with respect to some value/ideology. & \textbf{A:} We voted in favor of increasing data privacy regulations.

\textbf{B:} We are working on introducing very precise targeted advertising. & In $B$, one should know that targeted advertising usually relies on extensive user data for higher precision, which inherently conflicts with data privacy. Conversely, $A$ supports data privacy. Thus, it is an Indirect inconsistency. \\ \bottomrule
\end{tabular}
\caption{Typology of Inconsistencies}
\label{tab:taxonomy}
\end{table*}

Our task formulation is versatile and can be applied to various use cases, including detecting inconsistencies:
\begin{itemize}
\setlength\itemsep{0pt}  
    \setlength\parsep{0pt}   
\item Between a party's program or manifesto and its actual policies;
\item Across different platforms (e.g., Instagram vs. TikTok);
\item Among different party branches (e.g., federal state A vs. federal state B);
\item Within a party, among its members.
\end{itemize}

\subsection{Types of inconsistency}
\label{taxonomy}
It's crucial to clarify our definition of "inconsistency" given that numerous casual and philosophical interpretations exist. Inconsistency is not the same as \textbf{contradiction}, although contradictions can be viewed as a strong form of inconsistency, where both statements cannot be True at the same time \citep{Dowden_Inconsistency}.
In strict philosophical definitions, statements $A$ and $B$ are inconsistent with each other if both \textit{cannot} be True, and the truth of one entails the falsity of the other \citep{Wolfram_1989}. Our definition of inconsistency, however, allows for situations where the truth of one statement does \textit{not} necessarily entail the falsity of the other. It also permits that both statements could be true or false yet still be inconsistent. To capture the nuances of inconsistency, we conceptualize it as a spectrum with multiple levels (see Table \ref{tab:taxonomy}). We attach a visual version of our scale in Appendix \ref{vis_scale}.

\section{Relation to other work}
\label{sec:related_work_comparison}
Contradiction detection is a closely related area of research that focuses on detecting contradictions between a pair of texts \citep{de-marneffe-etal-2008-finding}. While previous studies have explored Contradiction detection in various domains, such as Finance \citep{Deuer2023ContradictionDI}, Medicine \citep{makhervaks-etal-2023-clinical}, contracts \citep{koreeda-manning-2021-contractnli-dataset}, or long documents \citep{yin-etal-2021-docnli, li-etal-2024-contradoc}, we are not aware of existing NLP research and resources on detecting inconsistencies in political domain. Our work aims to fill this gap by introducing a benchmark dataset consisting of real-world-grounded political statements. In Appendix \ref{contradictions_overview}, we additionally overview contradiction types used in other papers and map them to classification proposed by us.

\subsection{Inconsistency Detection Vs Fact-Checking}
Fact-checking evaluates the accuracy of statements made by politicians, journalists, and other public figures \citep{Graves_Amazeen_2019}. Only claims containing a \textbf{purported fact} can be fact-checked \citep{DAS2023103219}. For instance, an opinion such as “Green is the best color” can not be verified since its truthfulness will vary from person to person. Note that in our task, two statements that are both False can either be consistent or inconsistent, depending on their relationship to each other, but independent of an external truth value.

\subsection{Inconsistency Detection Vs Stance Detection}

Stance detection involves identifying an actor's position or attitude toward a specific target topic. Usual labels for attitude are \textit{\{Favor, Against, None\}} \citep{ALDAYEL2021102597}. The targets of Stance detection can be either pre-selected (e.g., given target "Climate change", detect the actor's stance on it) or open-ended, where targets are generated dynamically \citep{li-etal-2023-new}. Since Stance detection focuses on speakers' attitudes, it is not very well suited for detecting "neutral" inconsistencies between stated facts such as "combustion engines are contributing to global warming" and "global warming is solely caused by sun spots".

Moreover, targets in contradictions might be highly specific and hard to define, for example:

\textbf{A:} Political parties in Germany shouldn't influence the country's cultural life.

\textbf{B:} Theater ABC in Berlin staged a play financed by party A.

Supposedly, the target here is "Political parties in Germany influencing cultural life", which is hard to account for in advance if we use preset targets. On the other hand, generating such targets dynamically would create a huge universe of possible targets. 

\subsection{Inconsistency Detection Vs Inconsistency Detection in Summarization}

The objective of Inconsistency Detection in Summarization (IDS) is to, given the original text and its summary, detect whether the two are consistent or not \citep{laban-etal-2022-summac, fabbri-etal-2022-qafacteval, goyal-durrett-2020-evaluating}. Some works go beyond binary setting, extending the task to predict the types of factual errors \citep{chan-etal-2023-interpretable, pagnoni-etal-2021-understanding}. IDS setting might cover a limited number of contexts in our task: for example, when politicians are directly quoting the manifesto or other sources, one could treat their statement as a summary, and the source they are quoting as the document being summarized.

Still, in our setting, texts A and B are \textit{not} summaries of each other, but rather independent statements. This means that text B can contain novel information not present in A, and vice versa, while still being consistent. However, in the IDS, if a summary contains novel and accurate information not found in the original document, it is labeled as inconsistent \citep{laban-etal-2022-summac}.

\section{Sample generation}
\label{sec:sample_generation}
\begin{figure*}[t] 
    \centering
    \includegraphics[width=\textwidth]{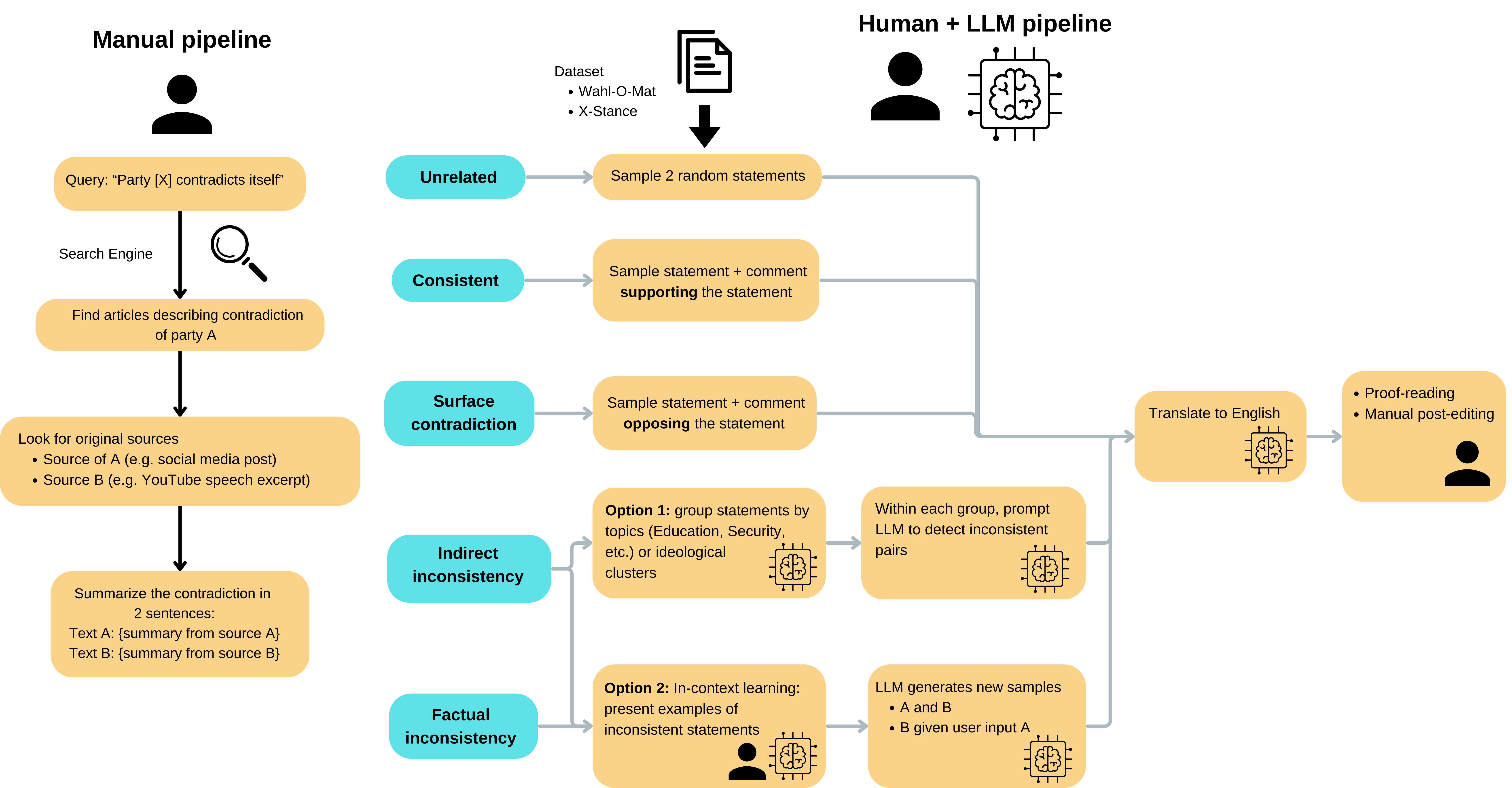} 
    \caption{Schematic description of the data annotation pipelines.}
    \label{fig:pipeline}
\end{figure*}

To create the dataset, we used several approaches: manual and human-LLM collaboration (see Figure \ref{fig:pipeline}). We use different pipelines to create samples for different classes, and focus on Inconsistent samples specifically, since we assume that Unrelated and Consistent classes will achieve higher consensus.

\subsubsection{Manual pipeline}
Our initial approach was to find examples of inconsistencies online. 
We searched on Google formulations such as:
"[party\_name] contradicts itself" or "[party\_name] being inconsistent" in German. 
Instead of party\_name we inserted names of German parties.  
Then we read through the articles we found, and either:
a) found the original sources of contradictions (e.g. YouTube speech excerpt, Facebook post, etc.)
and/or
b) summarized the main contradictions in the form "Text 1: A. Text 2: B". 
Samples obtained via manual pipeline are a minority in the dataset, as collecting them was highly time-consuming, and inconsistencies were relatively rare to find.

\subsubsection{Re-using existing datasets}

Wahl-O-Mat is a German tool for civic education\footnote{\url{https://www.bpb.de/themen/wahl-o-mat/}} that presents users with socio-political statements, such as "Businesses should be allowed to extend their Sunday opening hours". Users can express their stance as "Agree," "Disagree," or "Neutral." Political parties also declare their positions on these statements. The tool then matches users with the closest political parties. We utilized Qual-O-Mat dataset\footnote{\url{https://github.com/gockelhahn/qual-o-mat-data}, shared license-free} which contains statements and party's views over various states of Germany from 2002 to 2025 collected from the Wahl-O-Mat application. A notable portion of the statements (2220) also contains comments made by the parties.

We also used X-stance, a stance detection dataset composed of 67,000 comments by Swiss election candidates, addressing over 150 political issues \citep{vamvas2020xstancemultilingualmultitargetdataset}. This data is sourced from the Swiss voting advice application Smartvote\footnote{\url{https://www.smartvote.ch/de/home}, data shared under CC BY-NC 4.0 license}. In Figure \ref{fig:data_sample} in the Appendix, we provide a snapshot of both datasets. We also take inspiration from the Perspectrum dataset \citep{chen2018perspectives}, although we do not directly use it in our sample generation pipeline. 

\subsubsection{Sample generation for each class}
For \textit{Unrelated} class, we randomly sample N pairs of statements from Wahl-O-Mat dataset. For \textit{Consistent} samples, we randomly select N statements and pair each with a comment from the party that supports it.\footnote{Two authors of the paper manually checked every pair of samples. If we noticed they were correlated (e.g., they could potentially be classified as Consistent/Inconsistent), we re-sampled the statements.
} To maintain a standardized sample format, we limit our search to statements where the party’s comments are no longer than 160 characters. To generate \textit{Surface contradictions}, we randomly sample N Wahl-O-Mat statements and match them with N comments from the parties which expressed themselves against the statement. For \textit{Factual} and \textit{Indirect} inconsistency, we use several approaches: 
\begin{itemize}
    \item Use LLMs to a) group statements from both datasets into topics (e.g., "Education", "Fiscal Policy") and detect inconsistent texts within each topic; or b) review all presented statements and identify inconsistencies (used for Indirect type; see the prompt in Appendix \ref{detect_indirect}); 
    \item Provide LLMs with examples of inconsistent statements to a) generate new inconsistent samples from scratch; b) generate text B in response to input text A such that A and B are inconsistent with each other (used for both Factual and Indirect types; see the prompt in Appendix \ref{generate_incon});
    \item Manually create samples by either combining dataset samples or crafting new ones.
\end{itemize}

We combined manual approach and LLMs for translating, paraphrasing, and filtering samples. Specific names, geographic locations, and party names were removed to mitigate implicit bias. While one might argue that producing statements synthetically from scratch could introduce some leakage into the dataset, most samples were produced by resampling other datasets. Moreover, we manually post-edited all samples to ensure broad topic diversity relevant to Germany and Switzerland's political issues. 

While we aimed to produce a balanced number of statements that could potentially be labeled as specific classes, the final labels were assigned as a result of crowdsourced annotation (see Section \ref{dataset}). More details on sample generation can be found in Appendix \ref{appendix:sample_gen_details}.

\section{Dataset}
\label{dataset}

Our dataset consists of 698 annotated samples, of which 237 samples have at least one explanation for the chosen label. Table \ref{tab:label_counts} reflects a more detailed breakdown of samples by classes. We ensured that all samples from the main dataset were annotated by at least 5 humans. 

\begin{table}[ht]
    \centering
    \begin{tabular}{l r}
        \hline
        \textbf{Inconsistency Type} & \textbf{Count} \\
        \hline
        Surface contradiction & 183 \\
        Factual inconsistency & 122 \\
        Indirect inconsistency & 179 \\
        Consistent & 96 \\
        Unrelated & 118 \\
        \hline
    \end{tabular}
    \caption{Distribution of final annotations.}
    \label{tab:label_counts}
\end{table}

Since the perception of inconsistency in politics varies from person to person and might depend on their political views, we expect the Inconsistency detection task to be quite subjective. This reflects in the inter-annotator agreement scores: Krippendorff's alpha = \textbf{0.528 }for 5 classes (ordinal metric), and \textbf{0.507} for 3 classes (if we treat all inconsistency types as one Inconsistent class; nominal metric); more on the metric choice in Appendix \ref{interannotator_agreement}. A small subset of samples had more than 5 annotations; in this case, we randomly sampled 5 out of N samples to calculate the agreement. While the score is generally not high, it is common in other subjective tasks, such as judging toxicity of online discussions, emotions by facial expressions, or detecting mental manipulation \citep{wong-etal-2021-cross, wang-etal-2024-mentalmanip}.

To arrive at the ground truth, we take majority labels for each sample. In case of a tie, which happened in around 16\% of the samples, we randomly select one of the top-candidates with equal counts. Both the majority label and the individual labels are part of our dataset.

\subsection{Annotation}
\label{sec:annotation}
We recruited crowd-source workers through the Prolific platform\footnote{\url{https://www.prolific.com/}} and published surveys on Qualtrics\footnote{\url{https://www.qualtrics.com/}}. Participants were required to be fluent in English and possess at least an undergraduate degree. A basic understanding of politics and economics was recommended. First, we conducted several short rounds of surveys (7 to 10 questions). Since the task is non-trivial, we provided annotators practice sessions to calibrate the understanding of the task. Those who successfully passed comprehension checks were invited to a long study. In total we had 12 annotators evaluate 350 samples each, with one annotator's responses not included in the final dataset due to suspected inattentiveness. Participants were primarily educated in Social Sciences and Humanities and represented diverse political backgrounds and age groups. Compensation for short studies was at least £9.00 per hour, with the hourly rate increased to £12.21/hour for the long study in accordance with the National Living Wage in the UK for age 21 or over\footnote{\url{https://www.acas.org.uk/national-minimum-wage-entitlement}}. More details on annotators' demographics and annotation guidelines are available in the Appendix \ref{annotation_details}. We publish the self-reported political leaning together with the annotations to make it possible to evaluate a potential political bias in the annotations. 

\textbf{To prevent the usage of LLMs,} we used Prolific's Captcha verification feature in the beginning of the surveys. During the long studies, we sometimes asked annotators to briefly explain their reasoning to the question given before, without the option to look back at the previous question. These open-ended questions randomly appeared from 10 to 15\% of the time. Additionally, we disabled copy-paste options in the free-text field. 

\subsection{Repeated samples}
\begin{figure}[ht!]
  \centering
  \includegraphics[width=\linewidth]{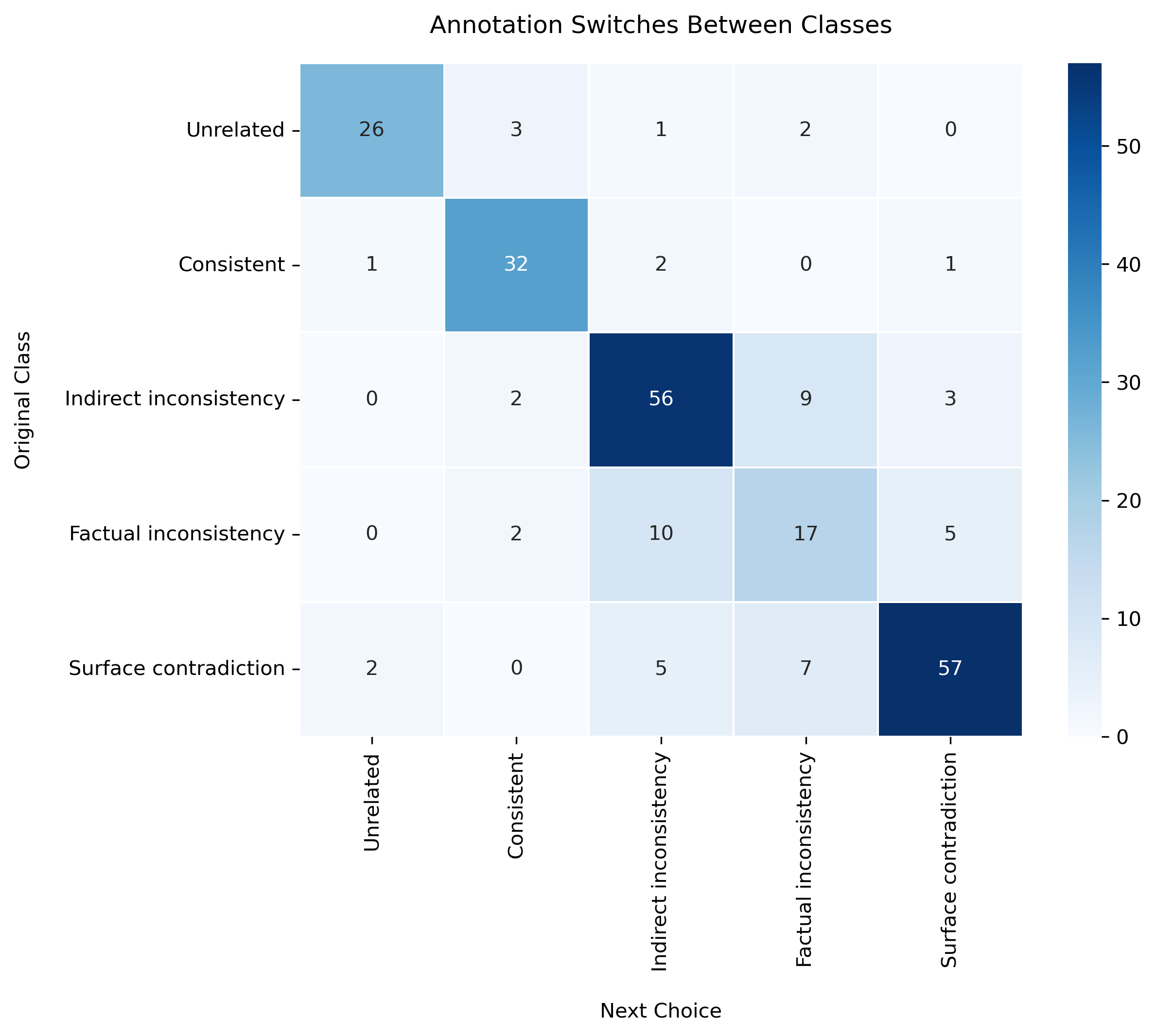}  
  \caption{Annotation switches within the same participant during repeated trials.}
  \label{fig:annotation_switches}
\end{figure}

Due to an error on our side, 190 samples were accidentally annotated repeatedly several times by the same participants. However, this allowed us to analyze consistency within the same annotator (see Figure \ref{fig:annotation_switches}). Most switches occurred within the "Inconsistent" class, with the most common case being a shift from Factual to Indirect Inconsistency. We only analyze the first two answers for each repeatedly annotated sample, since when labeling the same question for the third time, participants most likely have memorized their previous answers. Only the first answer per sample is included in the dataset.

\section{Model evaluation}
\label{sec:model_eval}

We evaluated four off-the-shelf models using the same instructions given to annotators, excluding visuals (see prompt in Appendix \ref{appendix:prompts_llms}). The following models were considered: \texttt{gpt-4-turbo-2024-04-09}, \texttt{gpt-3.5-turbo-instruct} (further referred to as ChatGPT-4 turbo and ChatGPT-3.5 turbo), LLaMA3.3 70B Instruct, and Llama3 8B Instruct. To obtain predictions for each model, we execute the same prompt five times to address prediction instability when identical prompts yield different labels.\footnote{An alternative way to obtain stable prediction would be to set the temperature to 0. We used default temperature and top-P settings.} We then select the majority label and resolve ties by randomly choosing among the top candidate labels. The costs are discussed in Appendix \ref{appendix:costs}.

\textbf{We first evaluate the performance of individual humans in predicting the majority label.} For each sample with $N$ human annotations, we used $N-1$ annotations to determine a majority label $L$, treating the remaining annotation as an individual human's prediction. We repeat this procedure $N$ times, obtaining $N$ pairs of "ground truth" and prediction for each sample. In cases of ties, we randomly selected one of the labels with the highest equal counts as the majority. Overall, this resulted in $N$×$M$ ground truth-prediction pairs, where $N$ is at least 5 and $M$ is 698, the total number of labeled samples. To evaluate model predictions of the majority label, we use the same setting, comparing model output with $N$ ground-truth labels per sample. We provide an illustration of the evaluation process in Appendix \ref{model_eval_visual}.

\begin{figure*}[ht]
    \centering
    \begin{subfigure}{0.45\textwidth}
        \centering
        \includegraphics[width=\linewidth]{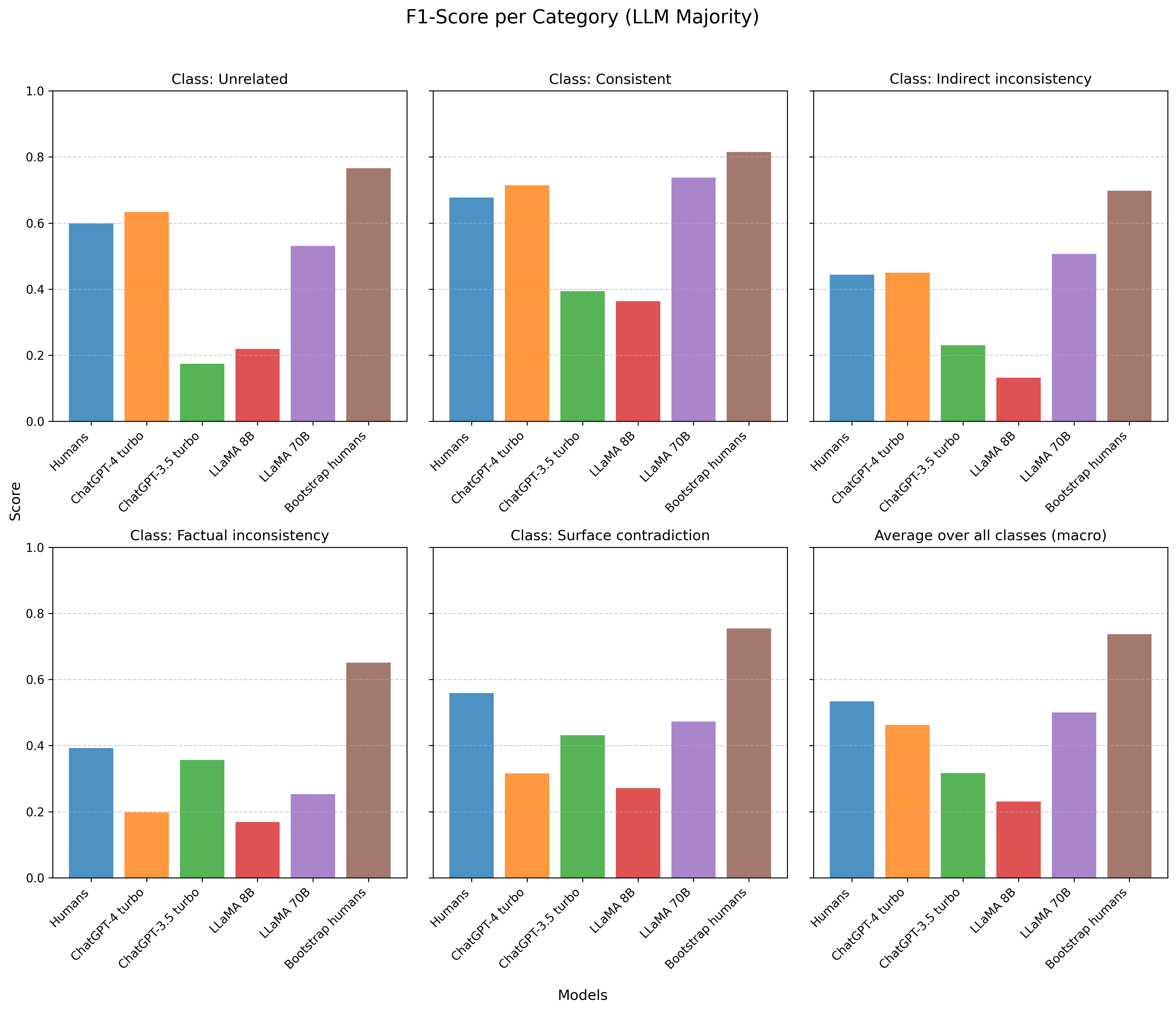}
        \caption{5 classes}
        \label{fig:f1_score_LLM_majority}
    \end{subfigure}
    \hfill
    \begin{subfigure}{0.45\textwidth}
        \centering
        \includegraphics[width=\linewidth]{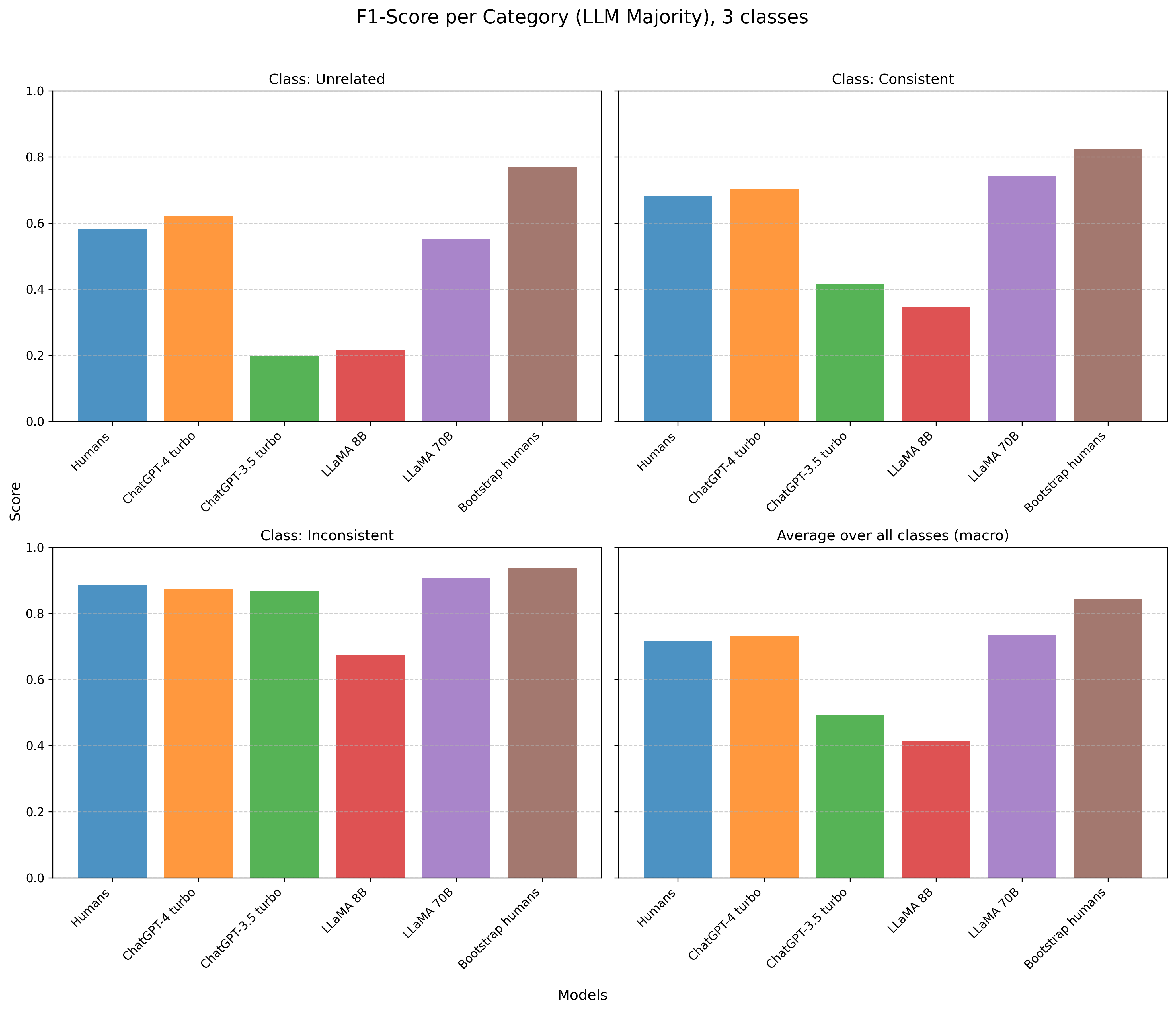}
        \caption{3 classes}
        \label{fig:f1_score_LLM_majority_3_classes}
    \end{subfigure}
    \caption{F1-score by model and class.}
    \label{fig:f1_score_LLM_majority_comparison}
\end{figure*}

\textbf{Due to the inherent labeling variation due to subjectivity, achieving 100\% accuracy on this task is impossible }and would indicate overfitting. To obtain a more realistic upper bound, we simulated a new set of annotators using bootstrapping with replacement. This allowed us to estimate how accurately annotations obtained on a different day could theoretically predict the ground truth established by our current human annotators. For each sample with $N$ annotations, we obtain $N$ tuples of size $(N-1)$. Then for each tuple, we generate 10 boostrapped versions by resampling it with replacement to simulate new annotations. We then determine the majority label $L$ for each of the $(N-1)$ x 10 tuples. In this process, we treat one held-out annotation as the ground truth and the 10 bootstrapped majority labels as predictions. Repeating this for each sample, we obtain $N$ x $10$ x $M$ tuples, where $N$ is at least 5 and $M$ is 698. 

\section{Results}

We analyze both 5-class and 3-class settings, where the latter treats all inconsistency types as a single "Inconsistent" class. 

\textbf{In the 3-class setting,} both humans and most models seem to approach the upper bound in predicting the general "Inconsistent" class (see Figure \ref{fig:f1_score_LLM_majority_comparison}b). 
\textbf{However, in the 5-class setting,} neither human annotators nor models reached this upper bound, indicating room for improvement (Figure \ref{fig:f1_score_LLM_majority_comparison}a). Interestingly, performance for most models and humans was lowest for Factual and Indirect inconsistency types, which were also the categories where annotators most frequently changed their minds during repeated trials (Figure \ref{fig:annotation_switches}). Based on F1-score, ChatGPT-4 turbo and LLaMA 70B showed the best overall performance in predicting the majority label, \textbf{sometimes even outperforming individual human annotators.}

\begin{table}[ht]
    \centering
    \renewcommand{\arraystretch}{1.1} 
    \setlength{\tabcolsep}{3pt} 
    \begin{tabular}{@{}l r r r@{}}
        \toprule
        \textbf{Model} & \textbf{Unrel.} & \textbf{Consist.} & \textbf{Inconsist.} \\ 
        \midrule
        ChatGPT-4 turbo & \textbf{0.548} & 0.662 & 0.619 \\  
        LLaMA 70B & 0.525 & \textbf{0.707} & \textbf{0.633} \\  
        Humans & 0.503 & 0.637 & 0.617 \\ 
        \hline
        Bootstrap humans & 0.727 & 0.798 & 0.786 \\  
        \bottomrule
    \end{tabular}
    \caption{Performance for 3 classes by MCC}
    \label{tab:comparing_metrics}
\end{table}

\begin{table}[ht]
    \centering
    \renewcommand{\arraystretch}{1.1} 
    \setlength{\tabcolsep}{3pt} 
    \begin{tabular}{@{}l r r r@{}}
        \toprule
        \textbf{Model} & \textbf{Indirect} & \textbf{Factual} & \textbf{Surface} \\ 
        \midrule
        ChatGPT-4 turbo & 0.174 & 0.183 & 0.328 \\  
        LLaMA 70B & \textbf{0.278} & 0.215 & 0.388 \\  
        Humans & 0.248 & \textbf{0.256} & \textbf{0.418} \\  
        \hline
        Bootstrap humans & 0.591 & 0.573 & 0.675 \\  
        \bottomrule
    \end{tabular}
    \caption{Performance for Inconsistent class by MCC}
    \label{tab:comparing_metrics_inconsist_only}
\end{table}

We compare the top models using the Matthews Correlation Coefficient (MCC), which has been shown to account for class imbalance more effectively than F1-score \citep{Chicco2020}. MCC ranges from -1 to 1, where 1 indicates perfect prediction and -1 indicates an inverse prediction. Tables \ref{tab:comparing_metrics} and  \ref{tab:comparing_metrics_inconsist_only} feature the results. LLaMA 70B shows a slight advantage over both ChatGPT-4 turbo and humans in predicting Consistent and general Inconsistent classes, while ChatGPT-4 turbo slightly leads in predicting the Unrelated class. LLaMA 70B also excels in Indirect inconsistency predictions, though it remains well below the upper bound. \textbf{Individual humans remained best in predicting Factual inconsistency and Surface contradiction types.}

\begin{figure}[ht!]
  \centering
  \includegraphics[width=\linewidth]{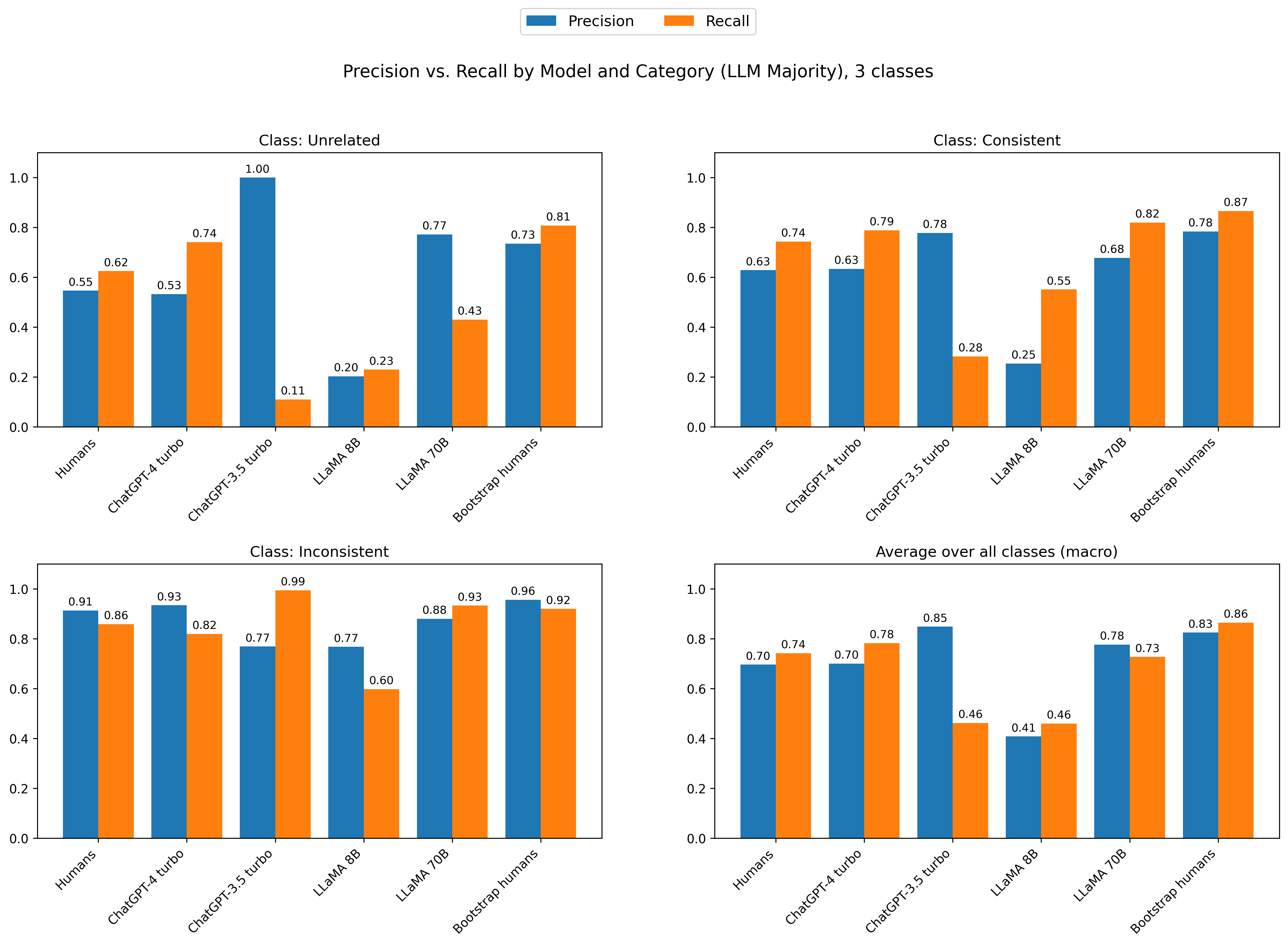}  
  \caption{Precision and Recall distribution for 3 classes.}
  \label{fig:precision_vs_recall_majority_3_classes}
\end{figure}

\begin{figure}[ht!]
  \centering
  \includegraphics[width=\linewidth]{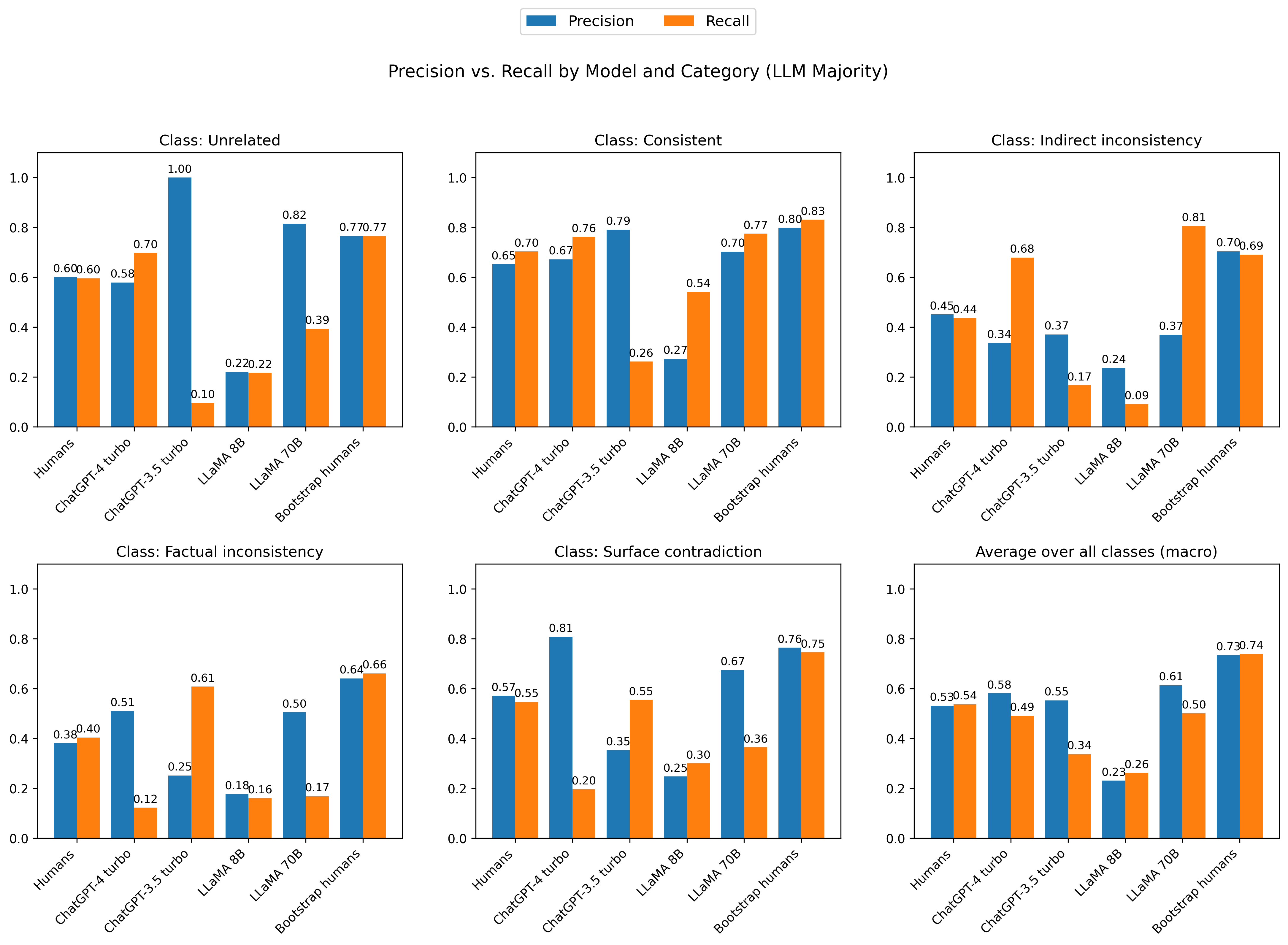}  
  \caption{Precision and Recall distribution for 5 classes.}
  \label{fig:precision_vs_recall_LLM_majority}
\end{figure}

To further dissect model performance, we analyze Precision vs. Recall scores (Figure \ref{fig:precision_vs_recall_majority_3_classes}). We observe that for humans, the difference between precision and recall is relatively modest across all classes. In contrast, for models the gap between precision and recall can be stark, especially in the 5-class setting. \textbf{This indicates that each model has a bias toward certain classes}, such as Indirect Inconsistency for LLaMA 70B or Factual Inconsistency for ChatGPT 3.5-turbo - Figure \ref{fig:precision_vs_recall_LLM_majority} shows high recall while notably lower precision for these classes. The underlying reasons for a certain class preference are not clear; further exploration of models' biases in perceiving inconsistency in politics is a potential future research direction (Section \ref{future_work}).

\section{Discussion and future directions}
\label{future_work}
This work and the corresponding dataset release lay the foundation for developing a system for detecting political inconsistencies in-the-wild. 
One can imagine a pipeline where parties' statements are routinely collected across different platforms (social media, parliament speeches, etc.) and checked for inconsistency with previous statements.
To build such a pipeline one has to address challenges related to not knowing which pairs of statement to check for consistency, in particular when dealing with documents of arbitrary length. Promising approaches include pre-filtering by topic to reduce the search space of potential inconsistencies. Absent of a filtering strategy, one would have to naively compare all pairs of statements, leading to scability challenges.

With such a system in place, a large number of empirical studies in computational social science would become feasible, such as comparisons of different parties, countries, and historic times.

\section{Limitations}
\label{sec:limitations}

There are several limitations we recognize in our work.

\textbf{First of all, we acknowledge that perception of inconsistency might be subjective} and depend on factors such as personal knowledge, education, political preferences, and even factors such as concentration and attentiveness while judging the statements. Because it is hard to strictly define degrees of inconsistency, we believe aligning models' understanding of inconsistency with human understanding is so important; humans as end users of the system should define what they care about and would find useful.

Given that our annotators were predominantly from Europe, it might have influenced the produced labels. Moreover, quite a large amount of samples to annotate (350 per annotator) could have influenced their attentiveness. We tried to address it by breaking down the surveys in two parts and randomly asking to explain the annotators' reasoning for the previous question, but the effects might still find place. 

\textbf{Second of all, we recognize geographical limitations of our dataset:} most of the samples were obtained from the context of German and Swiss politics, and the issues discussed in these regions might not be as relevant in other countries. \textbf{Thirdly, our samples are in English,} which might mean that current models would struggle with generalizing to other languages. We see potential improvement in these areas by collecting more diverse samples from countries with different political and economical contexts, and experimenting with generalization of the models to new examples of political inconsistencies.

\textbf{Another limitation is that our task takes pairs of texts as an input,} whereas inconsistency in politics is often also expressed by actions, such as voting for certain law projects. However, if such actions are documented in a form of text, we hope this doesn't pose a major limitation to performing Inconsistency detection.

\textbf{Lastly, the samples we present are short one-sentence statements}. We simplify the setting on purpose to make sure our annotators don't get lost in long documents and overlook potential inconsistencies. In the future, however, we aim to collect longer and noisier text samples and compare model performance.

\section*{Ethical Statement}
\label{sec:ethical_statement}
While focusing on beneficial use cases, we recognize that a misuse of Inconsistency detection systems is possible. For example, parties might use such tools to try and harass  their opponents. Moreover, when deploying such systems in real world, it would be important to make sure it treats all parties equally, without bias against or in support of any political views.

A broader question that remains is how important consistency in politics is to voters. Recent studies suggest that people tend to overlook political inconsistency as long as the current policy matches their preferences \citep{flip_flopping} and the effects of inconsistency are limited to certain circumstances  \citep{Karande2008WhenDA}. Thus,  candidates who have been inconsistent might be better off explaining the reasons for their change in position. 
 
Moreover, some studies suggest that politicians who are more averse to lying have lower reelection rates, indicating that honesty might not pay off in politics \citep{lying_aversion}. Still, we believe that due to information overload and scarce resources to monitor political parties' online presence attentively, many inconsistencies go unnoticed, and at least detecting them would be beneficial for both practical and research purposes.

Regarding the content of the dataset, we realize that it contains some samples that some people might find offensive  (for example, statements like "Headscarves for female teachers in public schools must be banned."). The statements presented in the dataset do not reflect personal views of the authors. The majority of the statements comes from the official voting assisting tools, such as Wahlomat and Smartvote. 

Regarding the annotation process, we received feedback that the task is rather intense, and therefore we broke down the main annotation into two parts. The annotators were given a week to annotate each part in their own tempo (not all questions at once), and were strongly encouraged to take a break when they start to get tired.

\section*{Acknowledgements}
Nursulu Sagimbayeva and Ingmar Weber are supported by funding from the Alexander von Humboldt Foundation and its founder, the Federal Ministry of Education and Research (Bundesministerium für Bildung und Forschung). We would like to thank our colleagues from the Interdisciplinary Institute for Societal Computing for their thoughtful feedback and suggestions.

We used AI tools to assist with tasks such as literature search (Elicit.ai\footnote{\url{https://elicit.com/}}) and minor coding and paraphrasing tasks (ChatGPT, Mistral). Every suggestion made by AI tools was verified by the authors.

\bibliography{aaai25}

\subsection{Ethics Checklist}

\begin{enumerate}

\item For most authors...
\begin{enumerate}
    \item  Would answering this research question advance science without violating social contracts, such as violating privacy norms, perpetuating unfair profiling, exacerbating the socio-economic divide, or implying disrespect to societies or cultures?
    \answerYes{Yes}
  \item Do your main claims in the abstract and introduction accurately reflect the paper's contributions and scope?
    \answerYes{Yes}
   \item Do you clarify how the proposed methodological approach is appropriate for the claims made? 
    \answerYes{Yes, we explain the model evaluation methodology and reasoning behind selected methods in section \ref{sec:model_eval}. Model evaluation}
   \item Do you clarify what are possible artifacts in the data used, given population-specific distributions?
    \answerYes{Yes, we talk about potential sampling bias in section \ref{sec:limitations}. Limitations}
  \item Did you describe the limitations of your work?
    \answerYes{Yes, section \ref{sec:limitations}. Limitations}
  \item Did you discuss any potential negative societal impacts of your work?
    \answerYes{Yes, section \ref{sec:ethical_statement} Ethical Statement}
      \item Did you discuss any potential misuse of your work?
    \answerYes{Yes, section \ref{sec:ethical_statement} Ethical Statement}
    \item Did you describe steps taken to prevent or mitigate potential negative outcomes of the research, such as data and model documentation, data anonymization, responsible release, access control, and the reproducibility of findings?
    \answerYes{Yes, section \ref{sec:sample_generation}. Sample generation (we anonymize names of politicians and parties)}
  \item Have you read the ethics review guidelines and ensured that your paper conforms to them?
    \answerYes{Yes}
\end{enumerate}

\item Additionally, if your study involves hypotheses testing...
\begin{enumerate}
  \item Did you clearly state the assumptions underlying all theoretical results?
    \answerNA{N/A}
  \item Have you provided justifications for all theoretical results?
    \answerNA{N/A}
  \item Did you discuss competing hypotheses or theories that might challenge or complement your theoretical results?
    \answerNA{N/A}
  \item Have you considered alternative mechanisms or explanations that might account for the same outcomes observed in your study?
    \answerNA{N/A}
  \item Did you address potential biases or limitations in your theoretical framework?
    \answerNA{N/A}
  \item Have you related your theoretical results to the existing literature in social science?
    \answerNA{N/A}
  \item Did you discuss the implications of your theoretical results for policy, practice, or further research in the social science domain?
    \answerNA{N/A}
\end{enumerate}

\item Additionally, if you are including theoretical proofs...
\begin{enumerate}
  \item Did you state the full set of assumptions of all theoretical results?
    \answerNA{N/A}
	\item Did you include complete proofs of all theoretical results?
    \answerNA{N/A}
\end{enumerate}

\item Additionally, if you ran machine learning experiments...
\begin{enumerate}
  \item Did you include the code, data, and instructions needed to reproduce the main experimental results (either in the supplemental material or as a URL)?
    \answerYes{Yes, we included prompts in Appendices \ref{appendix:sample_gen_details}, \ref{appendix:prompts_llms}, and we make our dataset and code publically available}
  \item Did you specify all the training details (e.g., data splits, hyperparameters, how they were chosen)?
    \answerNA{N/A}
     \item Did you report error bars (e.g., with respect to the random seed after running experiments multiple times)?
    \answerNo{No}
	\item Did you include the total amount of compute and the type of resources used (e.g., type of GPUs, internal cluster, or cloud provider)?
    \answerYes{Yes, Appendix \ref{appendix:costs}}
     \item Do you justify how the proposed evaluation is sufficient and appropriate to the claims made? 
    \answerYes{Yes, section \ref{sec:model_eval}. Model evaluation}
     \item Do you discuss what is ``the cost`` of misclassification and fault (in)tolerance?
    \answerNo{No}
  
\end{enumerate}

\item Additionally, if you are using existing assets (e.g., code, data, models) or curating/releasing new assets, \textbf{without compromising anonymity}...
\begin{enumerate}
  \item If your work uses existing assets, did you cite the creators?
    \answerYes{Yes, section \ref{sec:sample_generation}. Sample generation}
  \item Did you mention the license of the assets?
    \answerYes{Yes, section \ref{sec:sample_generation}. Sample generation}
  \item Did you include any new assets in the supplemental material or as a URL?
    \answerYes{Yes, we included a link to our dataset and code in the abstract}
  \item Did you discuss whether and how consent was obtained from people whose data you're using/curating?
    \answerNo{No, we didn’t explicitly ask for consent of annotators, however, we are not using the personal data of the annotators, but labels that they assigned to statements. We mentioned in the study description that we are conducting an Inconsistency detection task.}
  \item Did you discuss whether the data you are using/curating contains personally identifiable information or offensive content?
    \answerYes{Yes, we discussed potential content offensiveness in \ref{sec:ethical_statement} Ethical Statement, and we did not use personally identifiable data}
\item If you are curating or releasing new datasets, did you discuss how you intend to make your datasets FAIR?
\answerNo{No}
\item If you are curating or releasing new datasets, did you create a Datasheet for the Dataset? 
\answerYes{Yes, we addressed the questions it asks in our document in section \ref{dataset}. Dataset, but we did not literally fill out the datasheet}
\end{enumerate}

\item Additionally, if you used crowdsourcing or conducted research with human subjects, \textbf{without compromising anonymity}...
\begin{enumerate}
  \item Did you include the full text of instructions given to participants and screenshots?
    \answerYes{Yes, Appendix \ref{appendix:annotation_guidelines}. Annotation guidelines}
  \item Did you describe any potential participant risks, with mentions of Institutional Review Board (IRB) approvals?
    \answerNo{No, all tasks were designed to be low-risk, involving only non-sensitive political texts from public sources.}
  \item Did you include the estimated hourly wage paid to participants and the total amount spent on participant compensation?
    \answerYes{Yes, section \ref{sec:annotation}. Dataset. Annotation}
   \item Did you discuss how data is stored, shared, and deidentified?
   \answerNo{No, as we do not store personally identifiable participant data}
\end{enumerate}

\end{enumerate}

\section*{Appendix}
\appendix

\section{Original statements from German politics examples (Figure \ref{fig:statement_2})}
\label{orig_statements_politics}

\subsection{Statements from the Green party}

\textbf{Text 1:} Wir Grüne im Bundestag stehen für Frieden, Abrüstung, kooperative Sicherheit und eine Kultur der militärischen Zurückhaltung sowie eine Stärkung der Parlamentsrechte. Unsere Politik zielt darauf ab, Konflikte gar nicht erst entstehen zu lassen. Wir fordern, die zivile Krisenprävention ins Zentrum deutscher Außenpolitik zu stellen und sich engagiert für internationale Abrüstung und Rüstungskontrolle einzusetzen. Wir unterstützen das Recht jedes Landes auf Selbstverteidigung nach Artikel 51 der VN-Charta. Darüber hinaus lehnen wir Waffenlieferungen in Kriegs- und Krisengebiete ab. 

(Source: \url{https://www.gruene-bundestag.de/themen/sicherheitspolitik#:~:text=Dar%C3%BCber%20hinaus%20lehnen%20wir%20Waffenlieferungen%20in%20Kriegs-%20und%20Krisengebiete%20ab})

Note: the original text was edited at the party's website as the paper was being written.

\textbf{Text 2: }

\textbf{Question}: Frage: Also kein Zurückweichen etwa bei Waffenlieferungen an die Ukraine?

\textbf{Answer:} Außenministerin Annalena Baerbock: Genau. Denn wenn wir nicht helfen, den brutalen russischen Angriff zurückzudrängen, würden wir an noch mehr ukrainischen Orten Horror und Leid sehen. 

(Source: \url{https://www.auswaertiges-amt.de/de/newsroom/interview-aussenministerin-baerbock-sz/2557862})

\subsection{Statements from the AfD}

\textbf{Text 1:} 13.6 Landwirtschaft: Mehr Wettbewerb.
Weniger Subventionen

(Source: \url{https://www.afd.de/wp-content/uploads/2023/05/Programm_AfD_Online_.pdf})

\textbf{Text 2: }Die AfD steht an der Seite unserer Landwirte. [...] Wir fordern: Die Verdopplung der Agrardiesel-Rückerstattung. 

(Source: \url{https://www.afd.de/sofortprogramm-landwirtschaft/})

\section{Inconsistency type not included in the final scale}
\label{stereotypes_category}

One more category that we considered adding is Stereotypes violation/Expectation violation, where the inconsistency cannot be properly explained logically and relies purely on one's expectations and experience \citep{Dowden_Inconsistency}. Consider the following statements:
\begin{quote}
    \textbf{A: }I am a Democrat. \textbf{B:} I don’t like latte.
\end{quote}

There exists a stereotype that US Democrats tend to drink latte more often than Republicans \citep{liberals_latte}.
However, it is hard to explain logically or ideologically why not liking latte as a Democrat is inconsistent. 
We refrained from including such inconsistencies into our classification scale due to a high subjectivity of this class and lack of reliable sources of politics-related stereotypes. Hence, such cases should be labeled as "Unrelated" according to our taxonomy.

\section{Scale visualization}

\begin{figure*}[ht]
    \centering
    \includegraphics[width=1\textwidth]{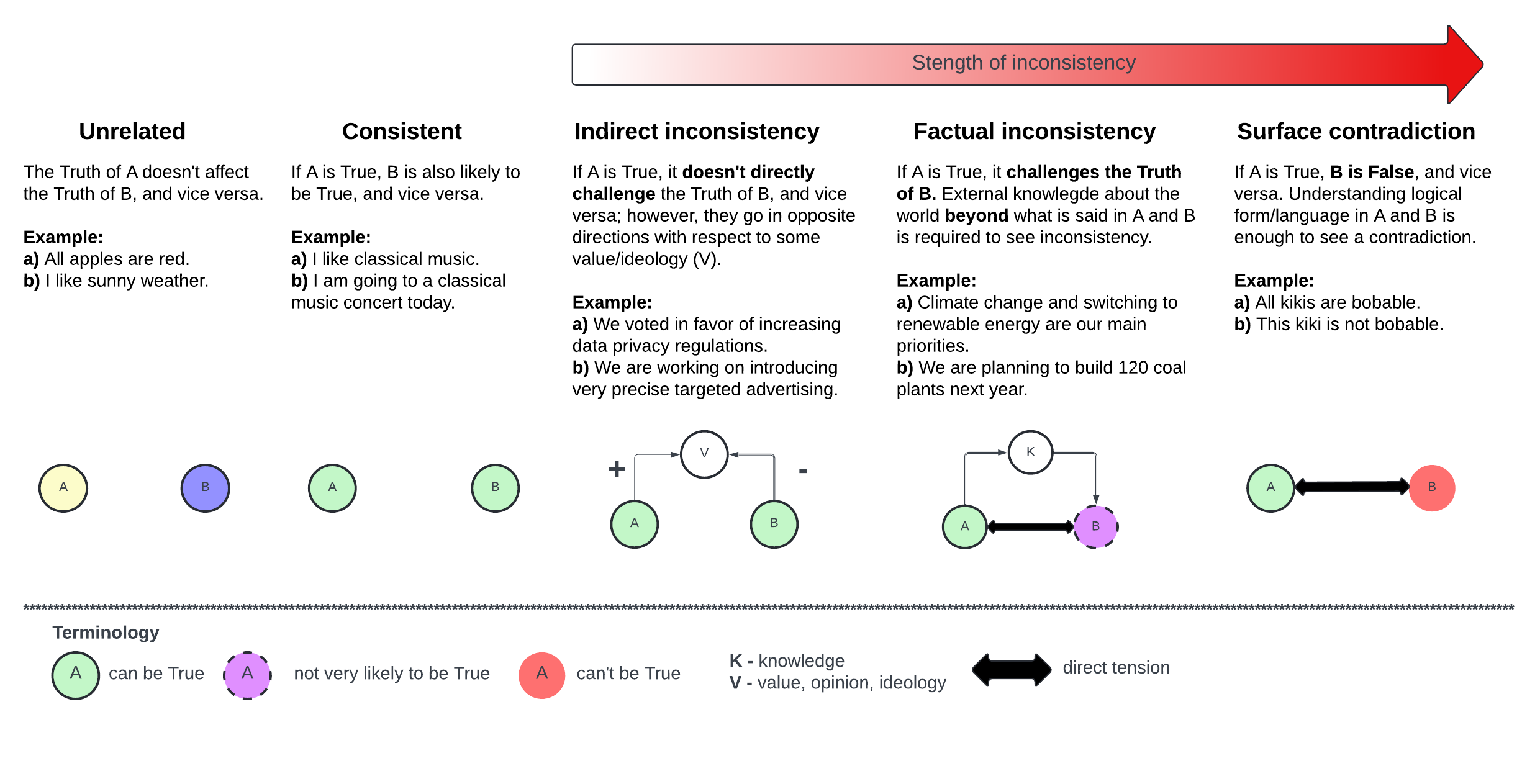} 
    \caption{Inconsistency scale visualization}
    \label{fig:scale_vis}
\end{figure*}

See Figure \ref{fig:scale_vis}.
\label{vis_scale}

\section{Overview of contradiction types}
\label{contradictions_overview}

See Figure \ref{tab:inconsistency_types_lit_review} for an overview of contradiction types in other literature and its relation to our taxonomy. Works marked with an asterisk * name their category differently, but based on their meaning, we categorize them as Factual inconsistency type.

There are some types of contradictions used in other papers that we don't include in our classification. 

\textit{Perspective / View / Opinion} and \textit{Emotion / Mood / Feeling} contradiction types used in \citep{li-etal-2024-contradoc} do not strictly fall into one of the inconsistency categories defined by us; they are rather orthogonal to our scale. For example, sentences:

A) We like the Blue party.

B) We hate the Blue party.

could be labeled as Surface contradiction, whereas sentences:

A)We are keen on forming a coalition with the Blue party.

B) We accuse John Smith of fraud in the elections. \textit{(given that John Smith is a leader of the Blue party).}

also signal shift in emotion/perspective, but would probably fall under Factual inconsistency category since noticing this inconsistency requires world knowledge about party leaders.

Another type that we do not include is\textit{ Structure contradiction} used in \citep{de-marneffe-etal-2008-finding, 10184012, Sepulveda_Torres_Bonet_Saquete_2021}. Structure contradictions are such that the structure of one of the sentences is not compatible with the other. This is achieved, for example, by interchanging named entities in the sentences: 

A) On January 26, 2014, Google acquired DeepMind.

B) On January 26, 2014, DeepMind acquired Google.

However, structure contradiction does not strictly fit one of the contradiction types that we adopt. For example, sentences:

A) Kiki wugged Boba.

B) Boba wugged Kiki.

would fall under Surface contradiction, whereas sentences:

A) Germany gave a large loan to Greece.

B) Greece gave a large loan to Germany.

would fall under the Factual inconsistency category, since noticing inconsistency requires knowledge about the economy and relations between these countries.

Moreover, sometimes change in structure doesn't lead to a contradiction: 

A) Merkel gave a present to Scholz. 

B) Scholz gave a present to Merkel.

Here, without having additional context, these two could happen simultaneously and are not incompatible.

We also did not include contradiction typology from domains that work with features beyond text, for example, rating-sentiment inconsistency \citep{SHAN2021113513}. 

\onecolumn
\begin{sidewaystable}
\centering
\scriptsize 
\begin{tabular}{|>{\centering\arraybackslash}p{2cm}|>{\centering\arraybackslash}p{2.5cm}|>{\centering\arraybackslash}p{4cm}|> 
{\centering\arraybackslash}p{6cm}|}

\hline
\textbf{Type (our scale)} & \textbf{Other names} & \textbf{Example}  & \textbf{Literature} \\ \hline
\multirow{4}{*}{Surface contradiction} & Logical & 1. All houses are \textit{green}. \newline
2. Sally's house is \textit{red}. &  \protect\citep{Lin_Zhang_2023} \\ \cline{2-4}
                       & Negation & 1) Sally donated her kidney.\newline 2) Sally \textit{never} donated her kidney. & \citep{de-marneffe-etal-2008-finding}, \citep{10184012}, \citep{li-etal-2024-contradoc}, \citep{Sepulveda_Torres_Bonet_Saquete_2021} \\ \cline{2-4}
                       & Numeric & 1) More
than \textit{50} civilians tragically died as a result of explosion.

2) The police found \textit{32} confirmed dead so far. & \citep{de-marneffe-etal-2008-finding}, \citep{10184012}, \citep{li-etal-2024-contradoc}, \citep{Sepulveda_Torres_Bonet_Saquete_2021}, \citep{Deuer2023ContradictionDI}\\ \cline{2-4}
                       & Antonyms & 1) Capital punishment is a \textit{catalyst} for more crime. 
2) Capital punishment is a \textit{deterrent} to
crime. & \citep{de-marneffe-etal-2008-finding}, \citep{10184012}, \citep{Sepulveda_Torres_Bonet_Saquete_2021} \\ \cline{2-4}
& Lexical & 1) The Canadian parliament’s Ethics Commission said
former immigration minister, Judy Sgro, \textit{did nothing
wrong} and her staff had put her into a conflict of interest.

2) The Canadian parliament’s Ethics
Commission \textit{accuses} Judy Sgro.
 & \citep{de-marneffe-etal-2008-finding}\\  \cline{2-4}
& Content & 1) She donated her kidney to a \textit{stranger}.

2) She donated her kidney to a \textit{close friend}. & \citep{li-etal-2024-contradoc}\\  \cline{2-4}
& Factive (Exaggeration) & 1) Isuzu and Volvo \textit{agree to create} a strategic alliance in heavy duty trucks. 

                        2) Isuzu and Volvo \textit{create} a strategic alliance in heavy duty trucks. & \citep{10184012}\\ \hline
\multirow{7}{*}{Factual inconsistency} & Factual & Abraham Lincoln is my mother. & \citep{de-marneffe-etal-2008-finding} *, \citep{huntsman2024prospectsinconsistencydetectionusing}*, \citep{aharoni-etal-2023-multilingual}, \citep{cohen-etal-2023-lm}, \citep{laban2023llmsfactualreasonersinsights}, \citep{xue2023rcotdetectingrectifyingfactual}, \citep{lattimer-etal-2023-fast}, \citep{xu-etal-2024-identifying}, \citep{makhervaks-etal-2023-clinical}, \citep{Lin_Zhang_2023} \\ \cline{2-4}
                        & World Knowledge & 1) Microsoft Israel, one of the first Microsoft branches outside the USA, was founded in 1989.
                        
                        2) Microsoft was established in 1989. & \citep{de-marneffe-etal-2008-finding} \\ \cline{2-4}
                        & Semantic & 1) On 14th of March, 2020, we increased our capital  by  offering  5,000  new  shares  during  a  seasoned equity offering.
                        
                        2) During  2020  we  did  not  increase  our  total amount  of equity  and  thus,  it  remained  unchanged at \$10,000,000. & \citep{Deuer2023ContradictionDI}\\ \cline{2-4}
                        & Causal & 1) I slam the door. 
                        
                        2) After I do that, the door opens. & \citep{li-etal-2024-contradoc} \\ \cline{2-4}
                        & Relation & 1) Jane and Tom are a married couple.
                        
                        2) Jane is Tom’s sister. & \citep{li-etal-2024-contradoc}\\ \hline
\multirow{1}{*}{Value inconsistency}  & Violation of expectations & I didn't attend the funeral, but I sent a nice letter saying I approved of it. & \citep{Lin_Zhang_2023} \\ \cline{2-4}
                        \hline
\end{tabular}
\caption{Types of Inconsistency in other literature.}
\label{tab:inconsistency_types_lit_review}
\end{sidewaystable}
\twocolumn

\section{Prompts for sample generation}

\subsubsection{Prompt for generating Factually inconsistent statement B for input A}
\label{generate_incon}

\begin{figure*}[t] 
    \centering
    \includegraphics[width=\textwidth]{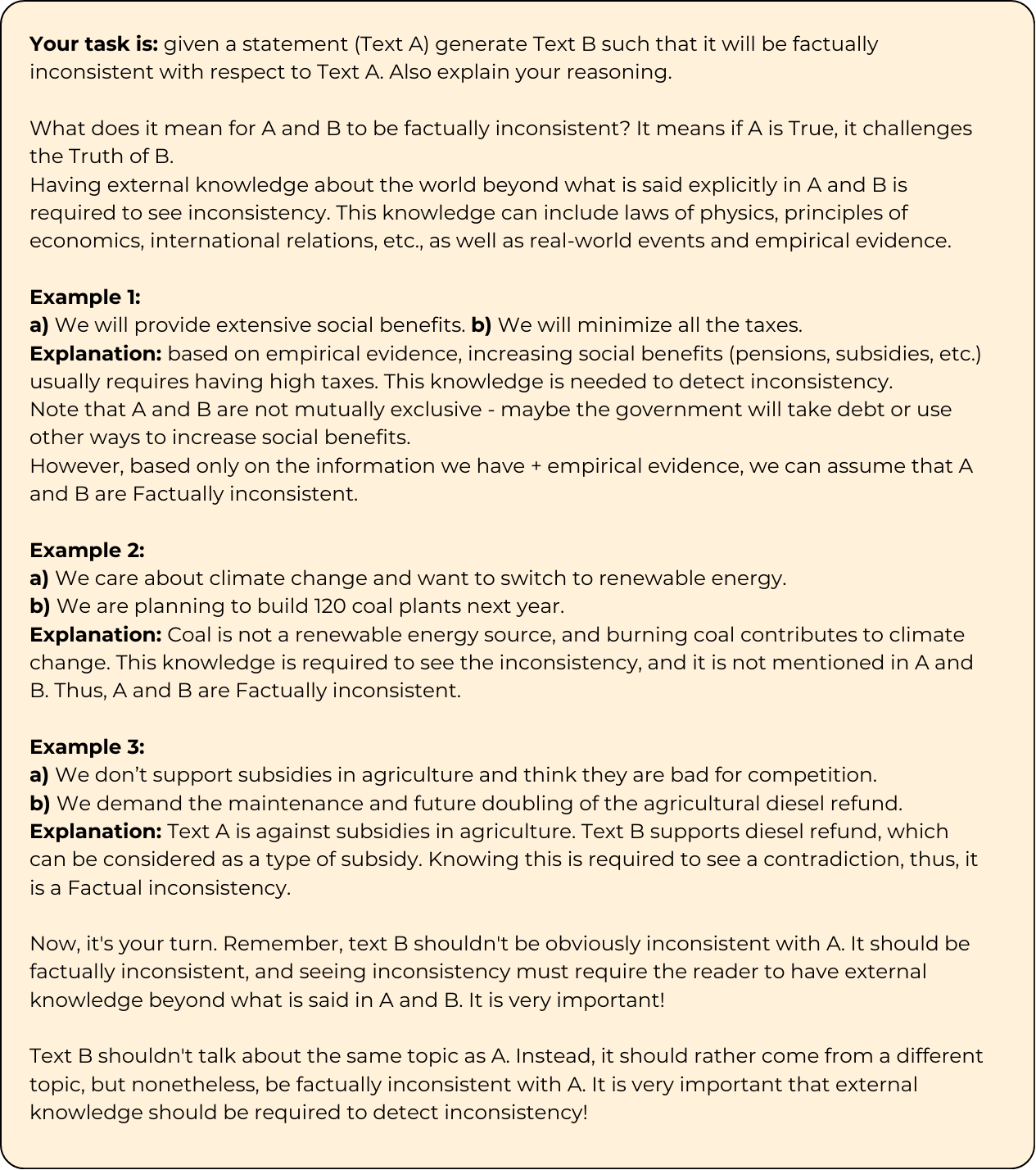} 
    \caption{Prompt for generating Factual inconsistencies).}
    \label{fig:prompt_factual}
\end{figure*}

See Figure \ref{fig:prompt_factual}.

\subsubsection{Prompt for detecting Indirect inconsistencies given a set of political statements}
\label{detect_indirect}

\begin{figure*}[hbtp] 
    \centering
    \includegraphics[width=\textwidth]{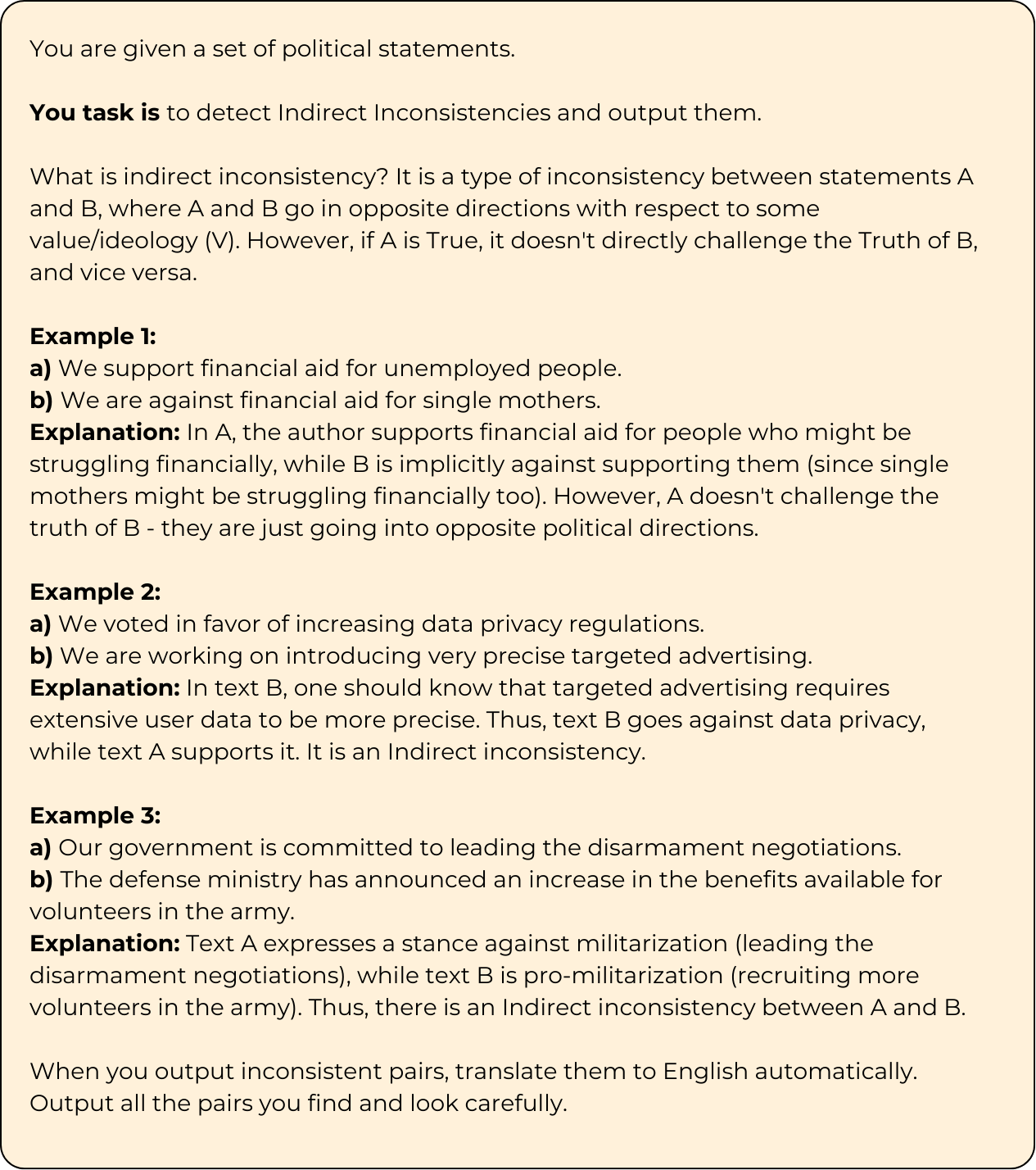} 
    \caption{Prompt for generating Indirect inconsistencies.}
    \label{fig:prompt_indirect}
\end{figure*}

See Figure \ref{fig:prompt_indirect}.

\section{Sample generation details}
\label{appendix:sample_gen_details}

Two authors of the paper manually checked every pair of samples. 

When randomly sampling Unrelated statements, we re-sampled them if we noticed they were correlated (e.g., they could potentially be classified as Consistent/Inconsistent).

During the random sampling of Unrelated statements, any samples that were correlated (e.g. they could be classified as Consistent or Inconsistent) were excluded and replaced through re-sampling.

For producing Factual and Indirect inconsistencies, we used \texttt{gpt-4-0613} and \texttt{gpt-4o-2024-11-20|} with temperature=1.0, top-P=1.0.

\subsection{Proportion of data produced by each method}
We estimate that around 50 samples were produced manually, by summarizing existing party contradictions or issues highlighted in other sources such as petitions from Change.org. Out of them, a subset of around 30 samples was used for annotation guidelines, prompting LLMs, and practice sessions for annotators. Thus, we did not include this subset in the final dataset to prevent data leakage. 

Around 80 samples were produced synthetically via prompting LLMs with examples of different Inconsistency classes. This low number is a result of stringent filtering since many LLM-produced samples were repetitive or did not seem complex enough for the Factual or Indirect Inconsistency classes.

The rest of the samples were produced by reusing the Wahl-O-Mat and X-stance datasets, and considerable manual labor was also invested in refining and rephrasing the statements.

 \section{Annotation details}
\label{annotation_details}

\subsection{Inter-annotator agreement}
\label{interannotator_agreement}
To measure agreement across five classes, we use Krippendorff's alpha with an ordinal metric, acknowledging that class order is important. For instance, misclassifying Surface contradiction as Factual inconsistency should be penalized less than misclassifying it as Consistent. For the three classes—Unrelated, Consistent, and Inconsistent — we applied Krippendorff's alpha with a nominal metric, as these categories are best treated as purely categorical without an inherent order. 

\subsection{Annotator Demographics}
\label{annotator_demographics}
Most participants resided in Europe, and some in the UK and the US. The majority of participants had obtained their education in Social Sciences and Humanities. To estimate possible bias in evaluating political statements, we asked participants to indicate their political standing. We obtained the following distribution of self-reported political views: Left-wing - 27\%, Rather left-leaning - 18\%, Moderate - 27\%, Rather right-leaning - 27\%. Age groups were distributed the following way: 18-24 - 4 people, 25-34 - 4 people, 35-44 - 2 people, 55-64 - 1 person. 

By trying to label some samples ourselves, we estimated that annotating each sample takes around 1 minute on average.

\subsection{Annotation guidelines}
\label{appendix:annotation_guidelines}
After providing instructions in short studies, we conducted a brief comprehension check consisting of four questions. Only participants who scored at least 2/4 were eligible for further consideration. 

To ensure a better understanding of the task before comprehension checks, we also conducted practice sessions, where participants could annotate a set of questions and see the answers and explanations we provided for them. The study was split into two parts, with a total of 350 samples per participant. We recruited 12 annotators for the main study. The answers from one annotator were discarded upon reviewing due to the explanations being too generic and possibly showing inattentiveness. Additionally, one participant chose not to participate in part 2 of the study, and was replaced by another annotator.

\begin{figure*}[htbp]  
  \centering
  \includegraphics[width=\textwidth, height=1\textheight, keepaspectratio]{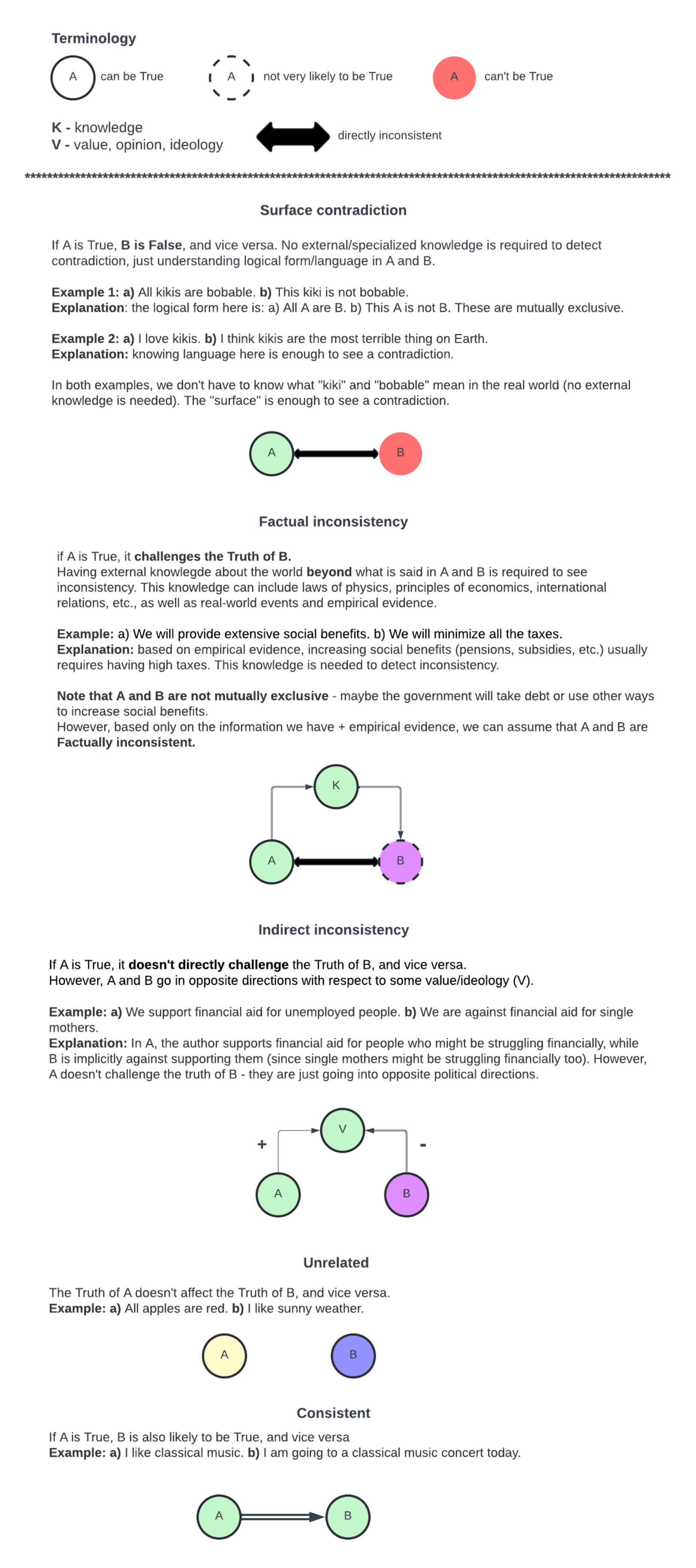}
  \caption{Instructions for annotators (Page 1)}\label{fig:annotator-1}
\end{figure*}

\begin{figure*}[htbp]  
  \centering
  \includegraphics[width=\textwidth, height=1\textheight, keepaspectratio]{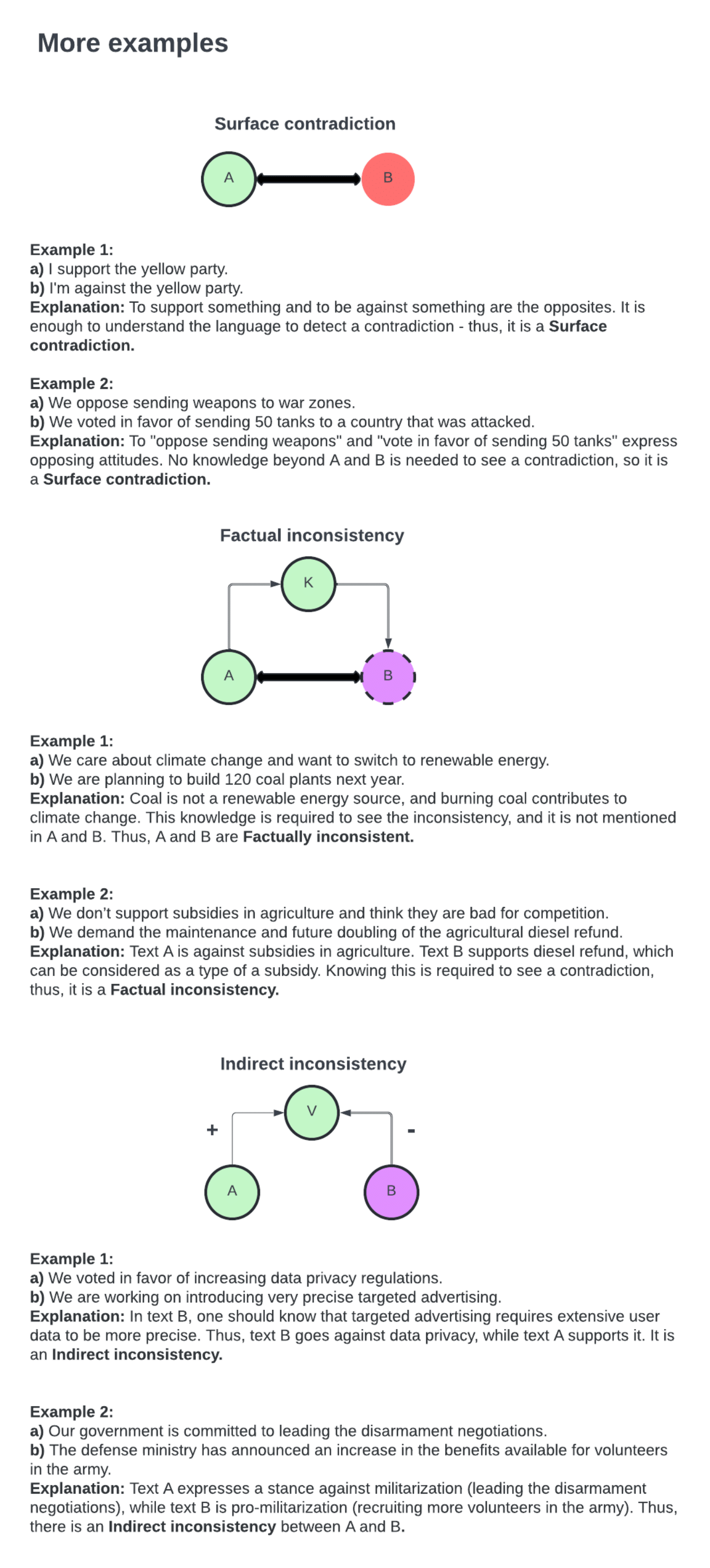}
  \caption{Instructions for annotators (Page 2)}\label{fig:annotator-3}
\end{figure*}

\begin{figure*}[htbp]  
  \centering
  \includegraphics[width=\textwidth, height=1\textheight, keepaspectratio]{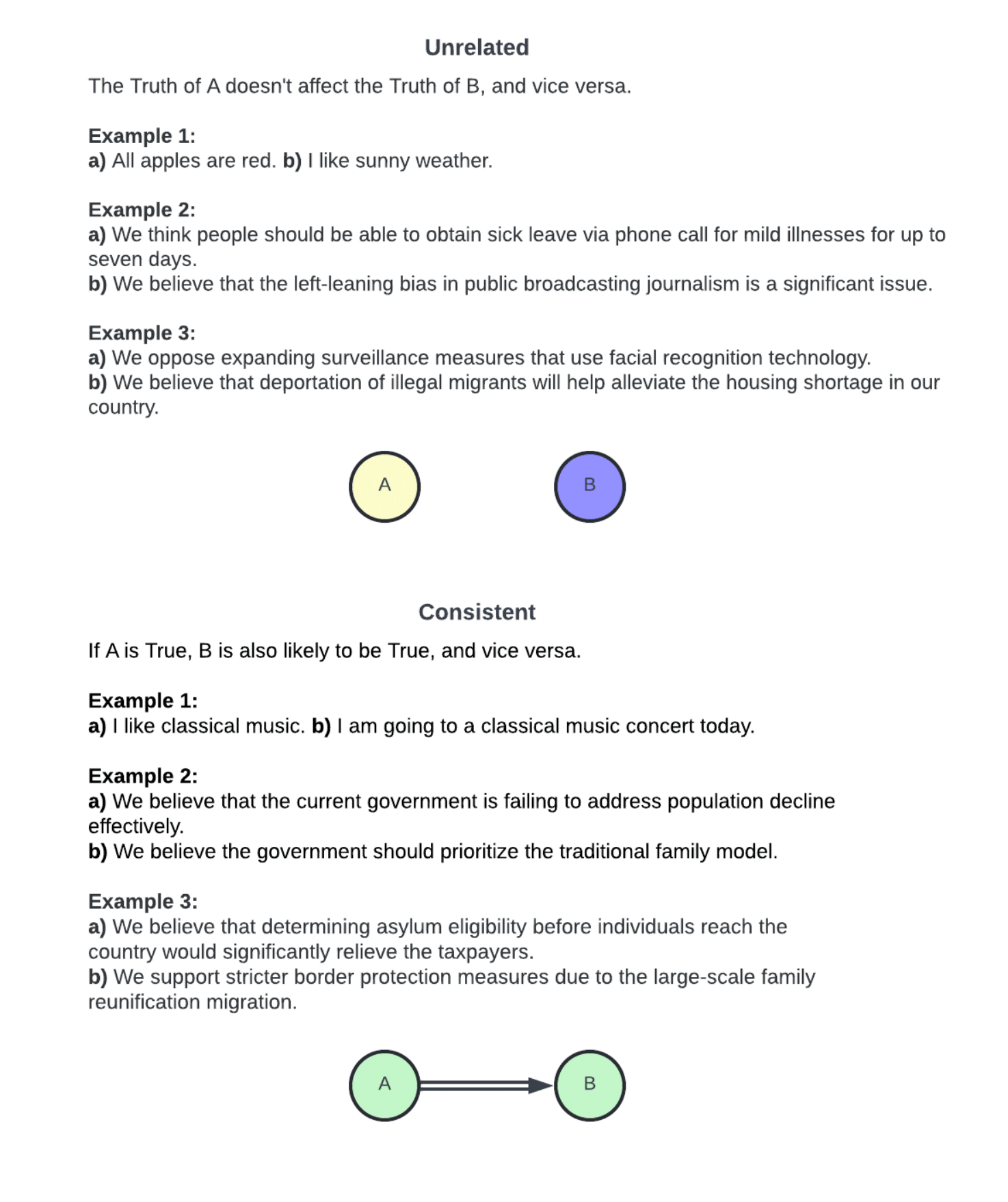}
  \caption{Instructions for annotators (Page 3)}\label{fig:annotator-4}
\end{figure*}

\begin{figure*}[htbp]  
  \centering
  \includegraphics[width=\textwidth, height=1\textheight, keepaspectratio]{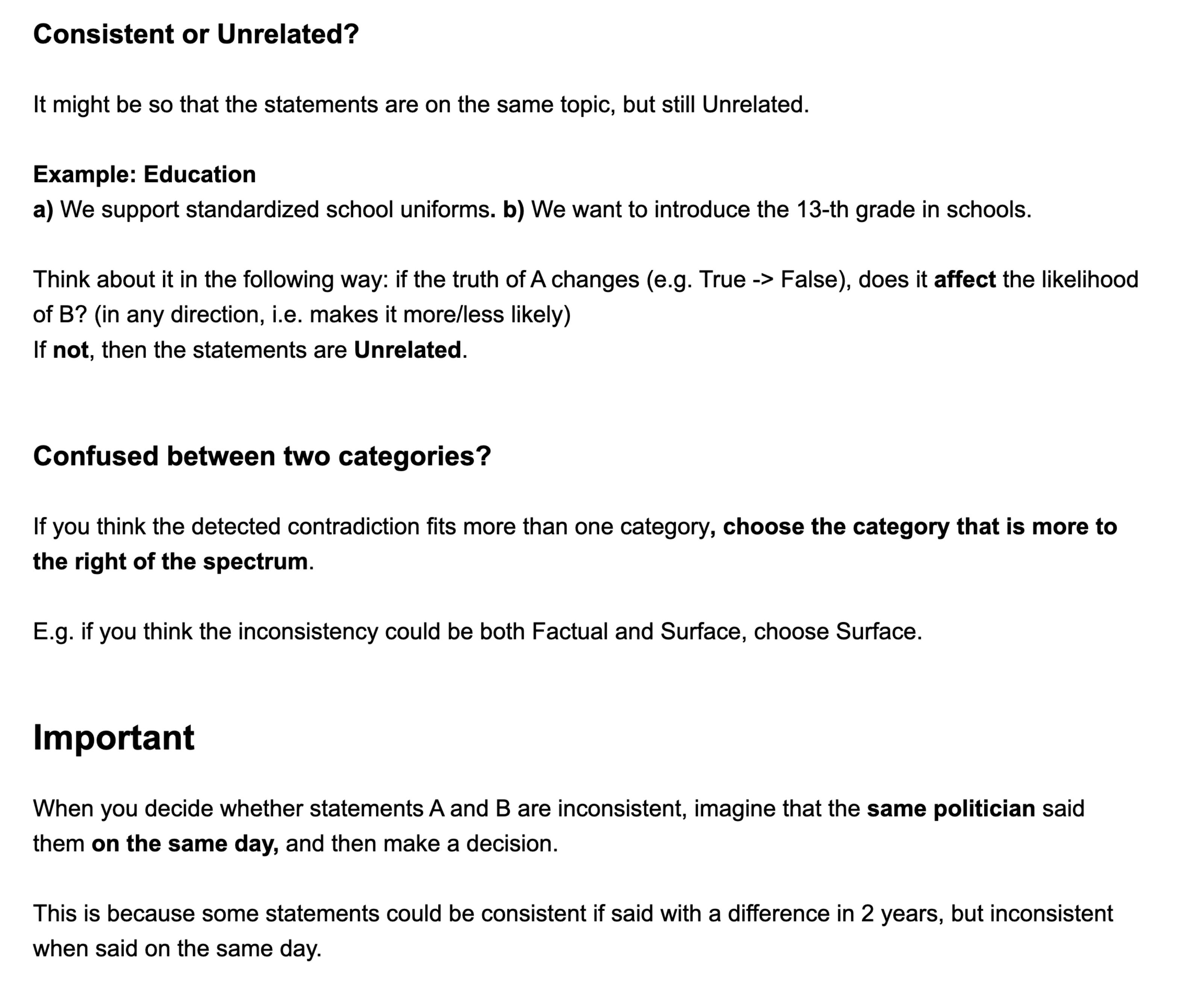}
  \caption{Instructions for annotators (Page 4)}\label{fig:annotator-5}
\end{figure*}

We attach screenshots from annotation guidelines in Figures \ref{fig:annotator-1}, \ref{fig:annotator-3}, \ref{fig:annotator-4}, \ref{fig:annotator-5}.

\clearpage  

\clearpage 

\clearpage 

\clearpage  

\subsection{Survey interface}

\begin{figure*}[ht]
  \centering
  \includegraphics[width=\linewidth]{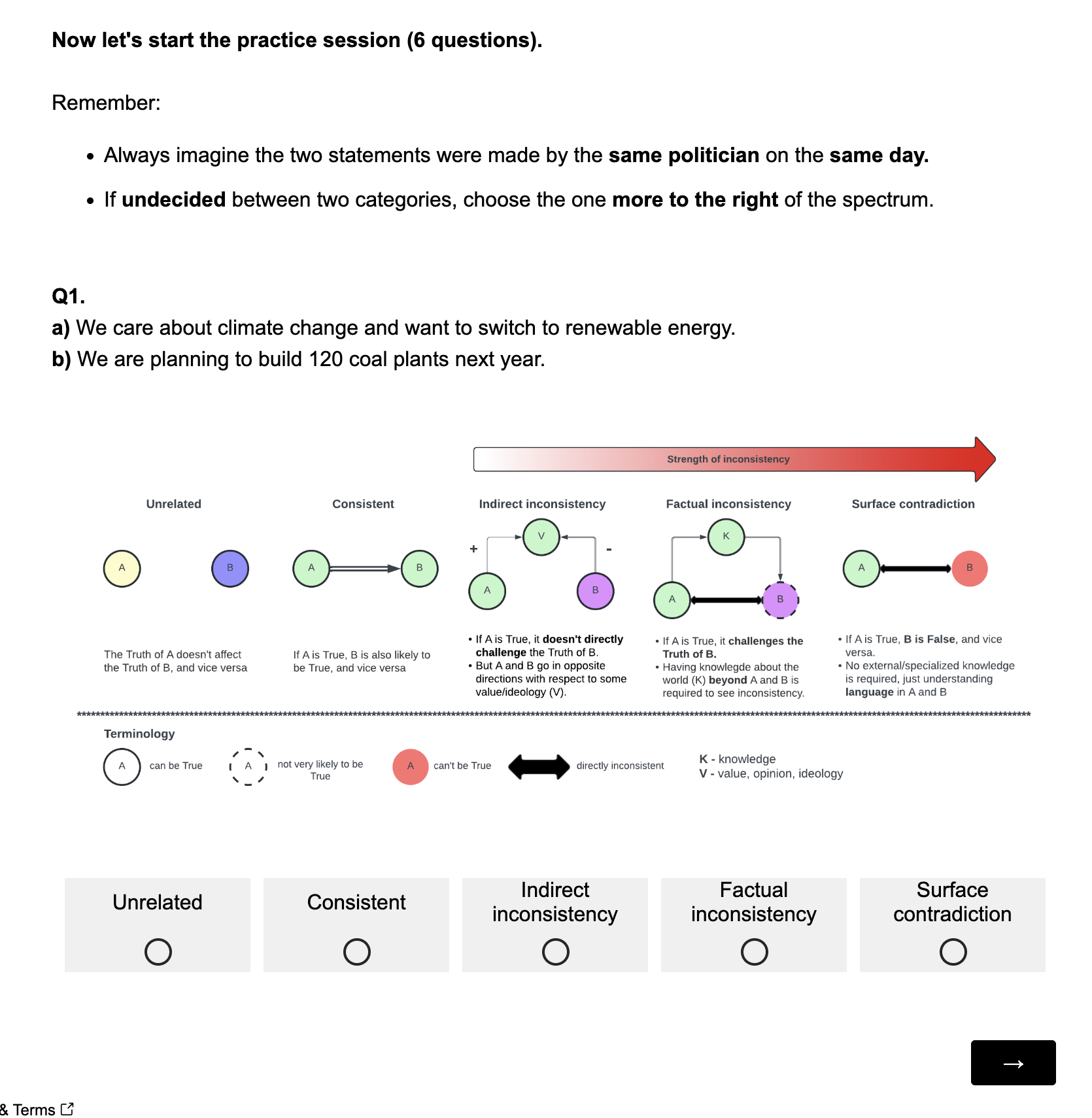} 
  \caption{Example from a practice session}
  \label{fig:survey_interface}
\end{figure*}

See Figure \ref{fig:survey_interface}.

\subsection{Samples from Wahl-O-Mat and X-stance}
\label{other_dataset_samples}

\begin{figure*}[t] =
    \centering
    \includegraphics[width=\textwidth]{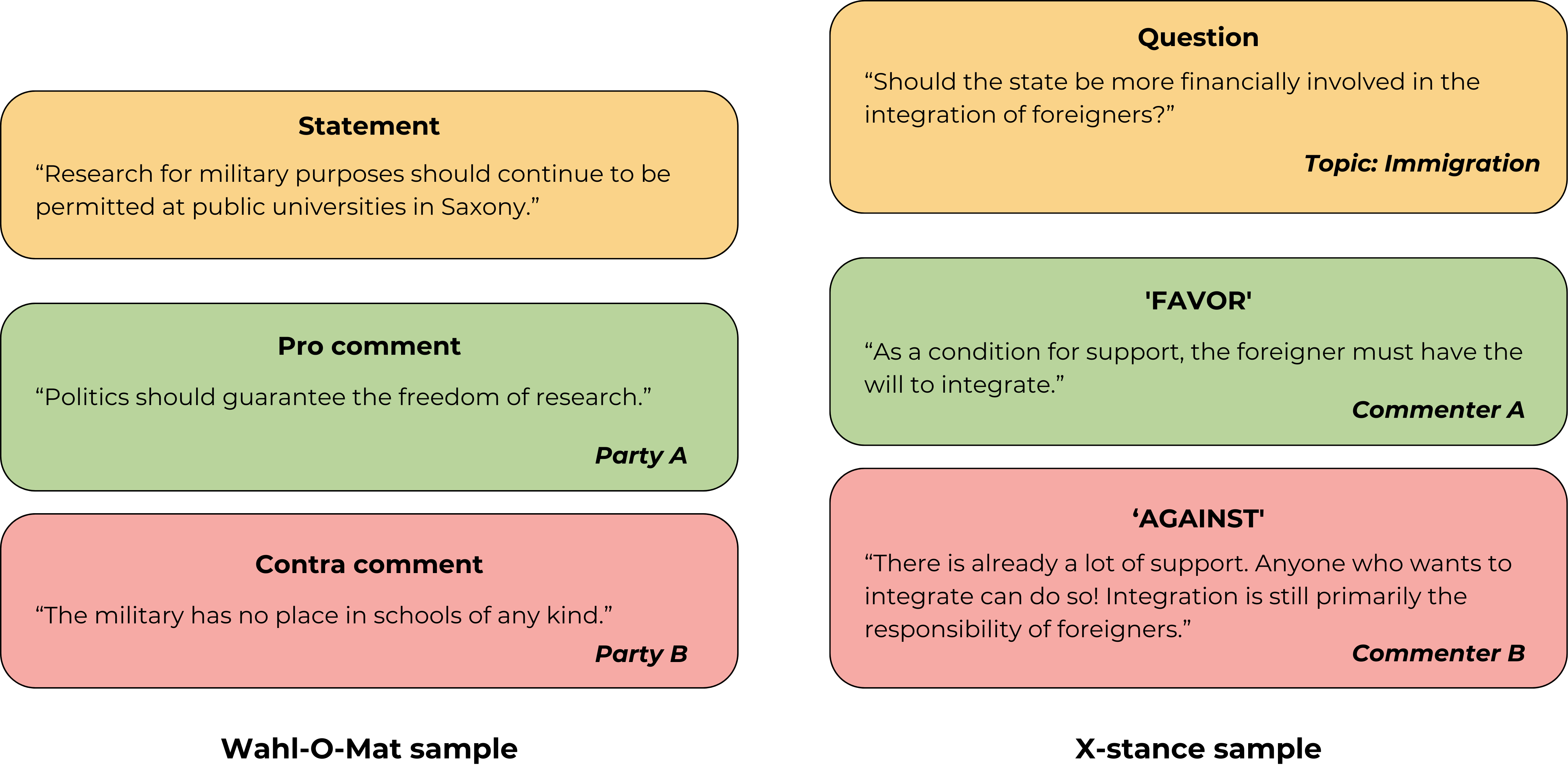} 
    \caption{Samples from two datasets (translated from German).}
    \label{fig:data_sample}
\end{figure*}

See Figure \ref{fig:data_sample}.

\section{Prompt for model evaluation}
\label{appendix:prompts_llms}
"You are a diligent annotator who is performing a crowdsourcing task. You will be given pairs of texts, which are statements made by a political party. Your task is to evaluate whether they are consistent with each other or not. Additionally, we will ask you to briefly explain your reasoning behind choosing the label.

Please familiarize yourself with the evaluation scale.
Evaluation scale 
(In growing order, the explanations will follow) 
Unrelated - Consistent - Indirect inconsistency - Factual inconsistency - Surface contradiction

Surface contradiction

If A is True, B is False, and vice versa. No external/specialized knowledge is required to detect contradiction, just understanding logical form/language in A and B.

Example 1: a) All kikis are bobable. b) This kiki is not bobable.
Explanation: the logical form here is: a) All A are B. b) This A is not B. These are mutually exclusive.

Example 2: a) I love kikis. b) I think kikis are the most terrible thing on Earth.
Explanation: knowing language here is enough to see a contradiction.

In both examples, we don't have to know what "kiki" and "bobable" mean in the real world (no external knowledge is needed). The "surface" is enough to see a contradiction. 

Example 3:
a) I support the yellow party.
b) I'm against the yellow party.

Explanation: To support something and to be against something are the opposites. It is enough to understand the language to detect a contradiction - thus, it is a Surface contradiction.

Example 4:
a) We oppose sending weapons to war zones.
b) We voted in favor of sending 50 tanks to a country that was attacked.

Explanation: To "oppose sending weapons" and "vote in favor of sending 50 tanks" express opposing attitudes. No knowledge beyond A and B is needed to see a contradiction, so it is a Surface contradiction.

Factual inconsistency

if A is True, it challenges the Truth of B.
Having external knowledge about the world beyond what is said in A and B is required to see inconsistency. This knowledge can include laws of physics, principles of economics, international relations, etc., as well as real-world events and empirical evidence.

Example 1: a) We will provide extensive social benefits. b) We will minimize all the taxes.

Explanation: based on empirical evidence, increasing social benefits (pensions, subsidies, etc.) usually requires having high taxes. This knowledge is needed to detect inconsistency.

Note that A and B are not mutually exclusive - maybe the government will take debt or use other ways to increase social benefits.
However, based only on the information we have + empirical evidence, we can assume that A and B are Factually inconsistent.

Example 2:
a) We care about climate change and want to switch to renewable energy.
b) We are planning to build 120 coal plants next year.
Explanation: Coal is not a renewable energy source, and burning coal contributes to climate change. This knowledge is required to see the inconsistency, and it is not mentioned in A and B. Thus, A and B are Factually inconsistent.

Example 3:
a) We don’t support subsidies in agriculture and think they are bad for competition.
b) We demand the maintenance and future doubling of the agricultural diesel refund.

Explanation: Text A is against subsidies in agriculture. Text B supports diesel refund, which can be considered as a type of a subsidy. Knowing this is required to see a contradiction, thus, it is a Factual inconsistency.

Indirect inconsistency

If A is True, it doesn't directly challenge the Truth of B, and vice versa.
However, A and B go in opposite directions with respect to some value/ideology (V).

Example 1: a) We support financial aid for unemployed people. b) We are against financial aid for single mothers.

Explanation: In A, the author supports financial aid for people who might be struggling financially, while B is implicitly against supporting them (since single mothers might be struggling financially too). However, A doesn't challenge the truth of B - they are just going in opposite political directions.

Example 2:
a) We voted in favor of increasing data privacy regulations.
b) We are working on introducing very precise targeted advertising.
Explanation: In text B, one should know that targeted advertising requires extensive user data to be more precise. Thus, text B goes against data privacy, while text A supports it. It is an Indirect inconsistency.

Example 3:
a) Our government is committed to leading the disarmament negotiations.
b) The defense ministry has announced an increase in the benefits available for volunteers in the army.
Explanation: Text A expresses a stance against militarization (leading the disarmament negotiations), while text B is pro-militarization (recruiting more volunteers in the army). Thus, there is an Indirect inconsistency between A and B.

Consistent

If A is True, B is also likely to be True, and vice versa.

Example 1:
a) I like classical music. b) I am going to a classical music concert today.

Example 2:
a) We believe that the current government is failing to address population decline effectively.
b) We believe the government should prioritize the traditional family model.

Example 3:
a) We believe that determining asylum eligibility before individuals reach the country would significantly relieve the taxpayers.
b) We support stricter border protection measures due to the large-scale family reunification migration.

Unrelated

The Truth of A doesn't affect the Truth of B, and vice versa.

Example 1:
a) All apples are red. b) I like sunny weather.

Example 2:
a) We think people should be able to obtain sick leave for mild illnesses for up to seven days.
b) We believe that the left-leaning bias in public broadcasting journalism is a significant issue.

Example 3:
a) We oppose expanding surveillance measures that use facial recognition technology.
b) We believe that the deportation of illegal migrants will help alleviate the housing shortage in our country.

Now you will be given a pair of statements. Evaluate them, and output only the label and the explanation in a format: 

Label: your label
Explanation: your explanation
********

Input texts: "

\section{Model running costs}
\label{appendix:costs}
We didn't keep precise track of our model evaluation budget.  However, based on the number of tokens, we estimate each complete run over 698 samples with OpenAI API for ChatGPT-4 turbo and ChatGPT-3.5 turbo together costed us around \$13. The prices might differ to some extent in reality. Moreover, under our prompt, the model outputs label \textbf{and its explanation,} which might have increased the costs. Below we list the prices relevant to the date:
\begin{itemize}
    \item \textbf{ChatGPT4-turbo:} \$10.00 / 1M input tokens  and \$30.00 / 1M output tokens. 
    \item \textbf{ChatGPT3.5-turbo instruct:} \$1.50 / 1M input tokens and \$2.00 / 1M output tokens.
\end{itemize}

Up-to-date prices can be looked up at OpenAI's official website: \url{https://openai.com/api/pricing/}.

For using LLaMA models InferenceAPI, we paid for the Pro account subscription on HuggingFace, which at the moment of writing the paper costed \$9.
Up-to-date prices can be looked up at HuggingFace's official website:
\url{https://huggingface.co/pricing}.

\section{Evaluation visualization (majority class prediction)}

\begin{figure*}[ht]
  \centering
  \includegraphics[width=\linewidth]{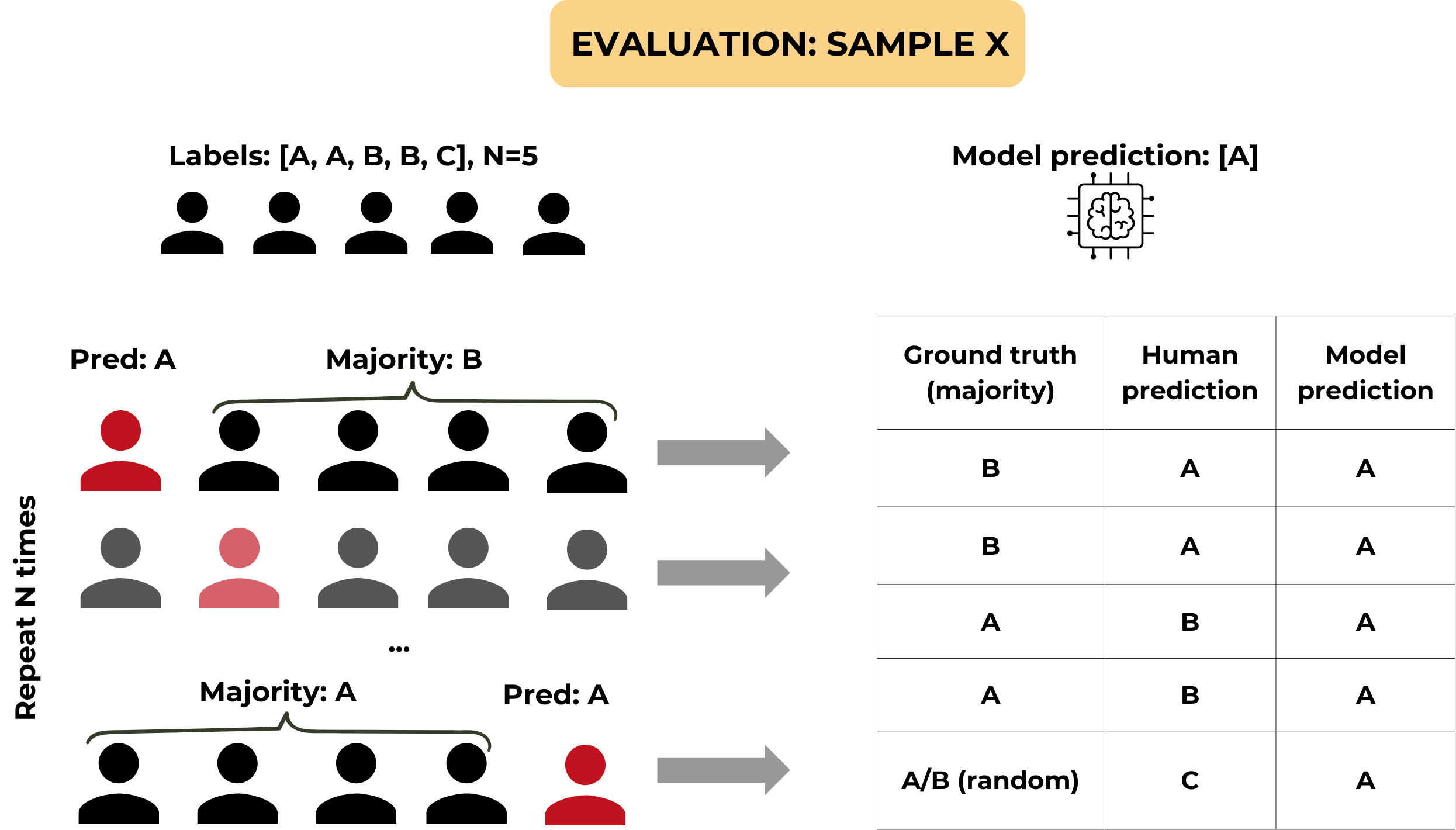}  
  \caption{A scheme for evaluating individual humans and models' prediction of the majority label.}
  \label{fig:eval_majority}
\end{figure*}

See Figure \ref{fig:eval_majority}.

\label{model_eval_visual}

\end{document}